\documentclass[10pt,journal,compsoc]{IEEEtran}

\usepackage{graphicx}
\usepackage[normalem]{ulem}
\usepackage{amsfonts,amssymb}
\useunder{\uline}{\ul}{}
\usepackage{amsmath,amssymb} 
\usepackage{color}
\usepackage{multirow}
\usepackage{booktabs}
\usepackage{array}
\usepackage{float}

\ifCLASSOPTIONcompsoc
  \usepackage[nocompress]{cite}
\else
  \usepackage{cite}
\fi

\begin{document}
%
\title{HMS-Net: Hierarchical Multi-scale Sparsity-invariant Network for Sparse Depth Completion}
%
%
%
%

\author{Zixuan Huang, Junming Fan, Shenggan Cheng, Shuai Yi, Xiaogang Wang,~\IEEEmembership{Senior Member,~IEEE,} \\Hongsheng Li,~\IEEEmembership{Member,~IEEE}
\IEEEcompsocitemizethanks{
\IEEEcompsocthanksitem Zixuan Huang is with Department of Computer Sciences, University of Wisconsin–Madison, United States
\IEEEcompsocthanksitem Junming Fan, Shenggan Cheng and Shuai Yi are with SenseTime Research, Beijing, China.\protect
\IEEEcompsocthanksitem Xiaogang Wang and Hongsheng Li are with the Department of Electronic Engineering at Chinese University of Hong Kong, Hong Kong, China.\protect
\IEEEcompsocthanksitem The first two authors contribute equally to the paper. They finished the
work while they were at SenseTime Research.\protect

\IEEEcompsocthanksitem E-mail: \{hsli, xgwang\}\@ee.cuhk.edu.hk}}

%
%

\markboth{Journal of \LaTeX\ Class Files,~Vol.~14, No.~8, August~2015}%
{Shell \MakeLowercase{\textit{et al.}}: Bare Advanced Demo of IEEEtran.cls for IEEE Computer Society Journals}
%



\IEEEtitleabstractindextext{%
\begin{abstract}
Dense depth cues are important and have wide applications in various computer vision tasks. In autonomous driving, LIDAR sensors are adopted to acquire depth measurements around the vehicle to perceive the surrounding environments. However, depth maps obtained by LIDAR are generally sparse because of its hardware limitation. The task of depth completion attracts increasing attention, which aims at generating a dense depth map from an input sparse depth map. 
To effectively utilize multi-scale features, we propose three novel sparsity-invariant operations, based on which, a sparsity-invariant multi-scale encoder-decoder network (HMS-Net) for handling sparse inputs and sparse feature maps is also proposed. Additional RGB features could be incorporated to further improve the depth completion performance. Our extensive experiments and component analysis on two public benchmarks, KITTI depth completion benchmark and NYU-depth-v2 dataset, demonstrate the effectiveness of the proposed approach. As of Aug. 12th, 2018, on KITTI depth completion leaderboard, our proposed model without RGB guidance ranks \textbf{1st} among all peer-reviewed methods without using RGB information, and our model with RGB guidance ranks \textbf{2nd} among all RGB-guided methods.
\end{abstract}

}

\maketitle

\IEEEdisplaynontitleabstractindextext

%
\IEEEpeerreviewmaketitle

\ifCLASSOPTIONcompsoc
\IEEEraisesectionheading{\section{Introduction}\label{sec:introduction}}
\else
\section{Introduction}
\label{sec:introduction}
\fi

%
%
%
%
\IEEEPARstart{D}{peth} completion, aiming at generating a dense depth map from the input sparse depth map, is an important task for computer vision and robotics. In Fig. \ref{fig:intro} (a), (b), (e), we show one example input sparse depth map, its corresponding RGB image, and the depth completion result by our proposed method. Because of the limitation of current LIDAR sensors, the inputs of depth completion are generally sparse. For instance, the \$100,000 Velodyne HDL-64E has only a vertical resolution of $\sim 0.4^{\circ}$ and an azimuth  angular resolution of $0.08^{\circ}$. It generates sparse depth maps, which might be insufficient for many real-world applications. Depth completion algorithms could estimate dense depth maps from sparse inputs and has great pontential in practice. With an accurate depth completion algorithm, many high-level vision tasks, such as semantic segmentation, 3D object detection, visual odometry and SLAM with 3D point clouds, could be solved more effectively. Therefore, it becomes a hot research topic for self-driving cars and UAVs, and is listed as one of the ranked tasks in the KITTI benchmark \cite{Uhrig2017THREEDV}.

Many different methods have been proposed for depth completion, which could be generally categorized into learning-based \cite{Uhrig2017THREEDV,liao2017parse, ma2017sparse,chodosh2018deep} and non-learning-based methods\cite{ku2018defense,nadaraya1964estimating,barron2016fast}. Non-learning-based approaches generate dense depth maps from sparse inputs based on hand-crafted rules. Therefore, the outputs of these algorithms are generated based on assumed prior by humans. As a result, they are not robust enough to sensor noises and are usually specifically designed for certain datasets. In addition, most non-learning-based methods ignore the correlations among sparse input depth points and might result in inaccurate object boundaries. An example of errors by a non-learning-based method \cite{ku2018defense} is shown in Fig. \ref{fig:intro}(e). The noises in the white box are not removed at all, and boundaries of the cars and trees in the yellow box are inaccurate.

\begin{figure}
	\centering
	\footnotesize
	\begin{tabular}{c@{\hspace{-3.5mm}}c}
		& \includegraphics[scale=0.35]{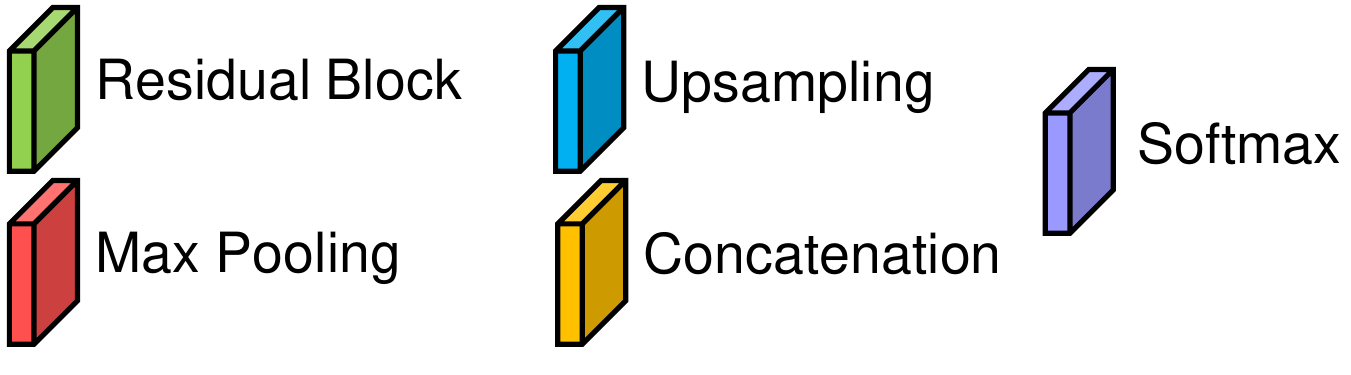} \\
		& \includegraphics[scale=0.35]{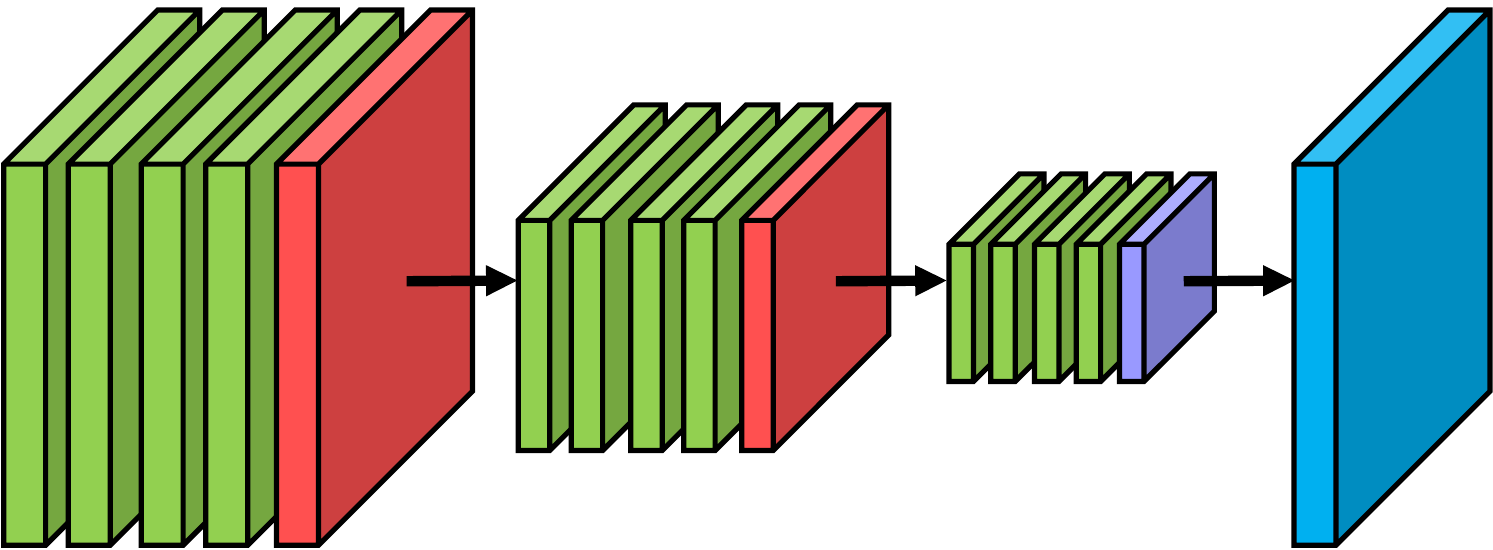} \\
		& (a) CNN with sparsity-invariant convolution \textbf{only}\\
		& \includegraphics[scale=0.35]{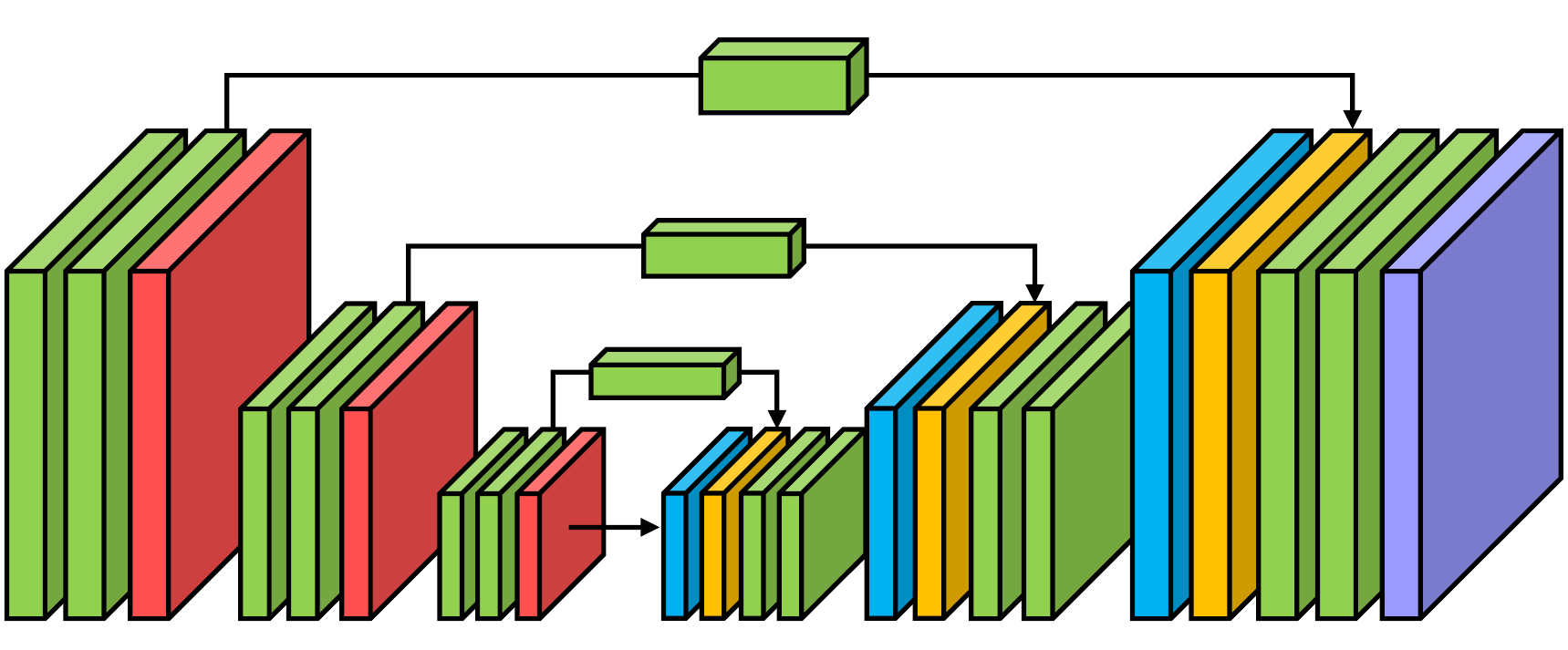}\\
		& (b) Proposed \textbf{Sparsity-invariant Encoder-decoder} Network
	\end{tabular}
	\caption{(a) CNN with only sparsity-invariant convolution could only gradually downsample feature maps, which loses much resolution at later stages. (b) Our proposed encoder-decoder network with novel sparsity-invariant operations could effectively fuse multi-scale features from different layers for depth completion.}
	\label{fig:convnet_encoder_decoder}
\end{figure}

\begin{figure*}[t]
\centering
\begin{tabular}{c c}
\includegraphics[height=2.25cm]{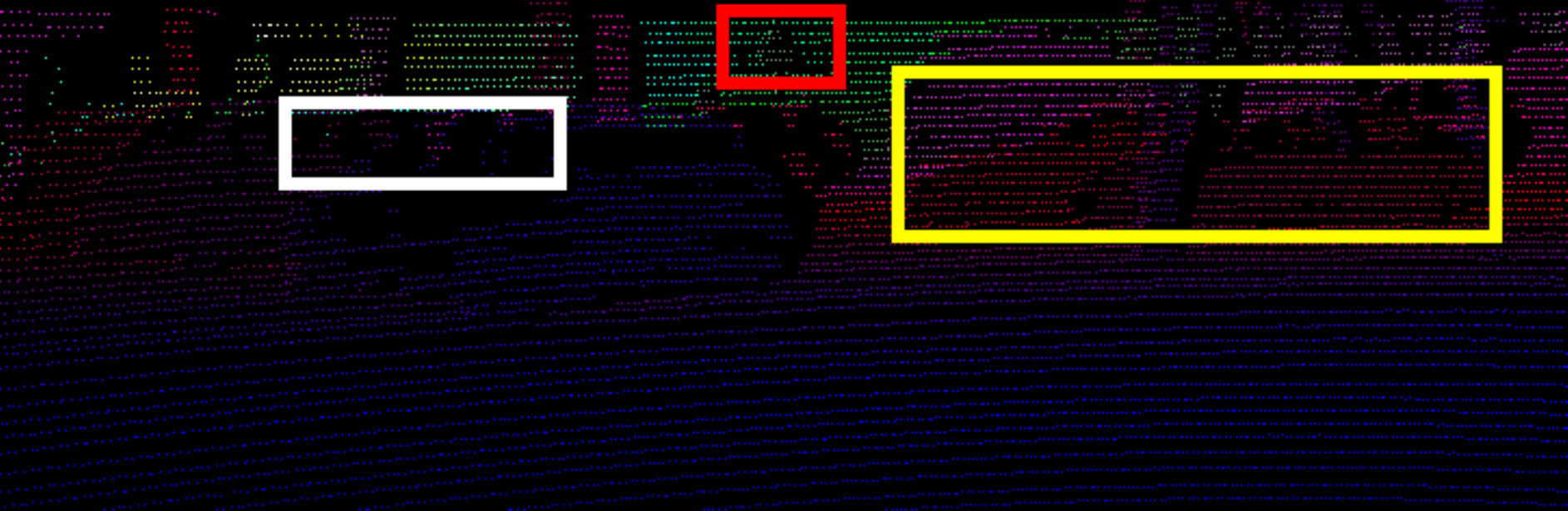} & \includegraphics[height=2.25cm]{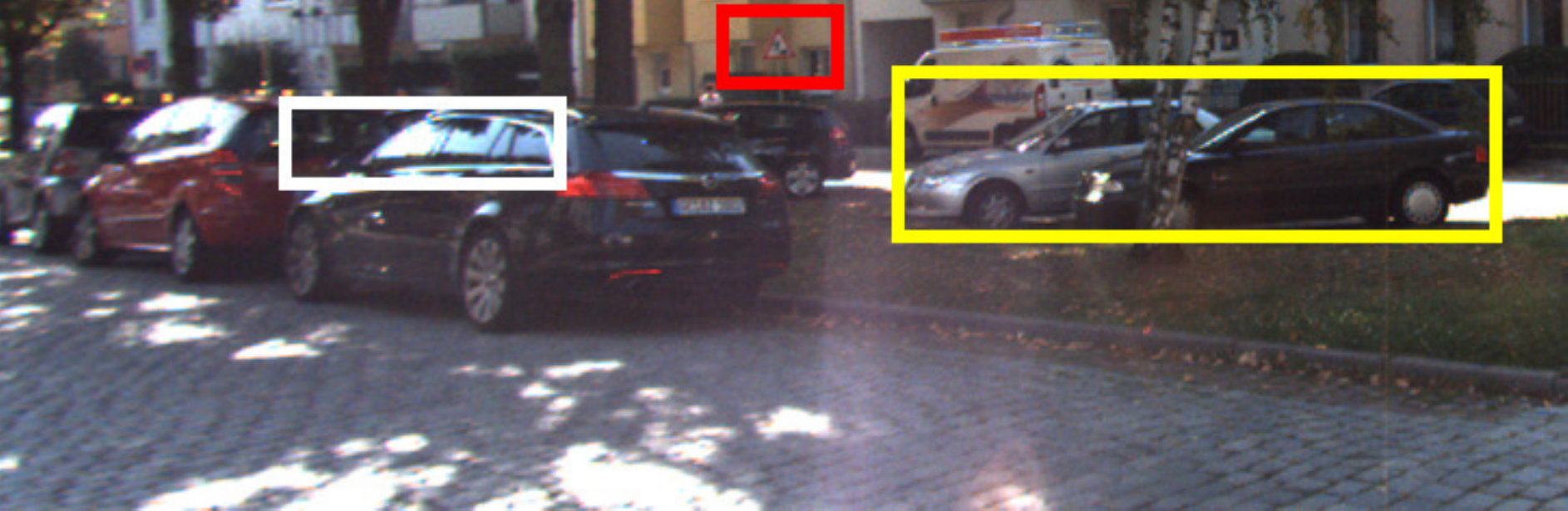} \\ 
(a) Input Sparse Depth Map & (b) Image \\ 
\includegraphics[height=2.25cm]{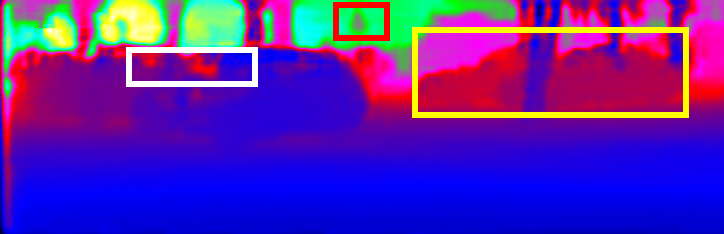} & \includegraphics[height=2.25cm]{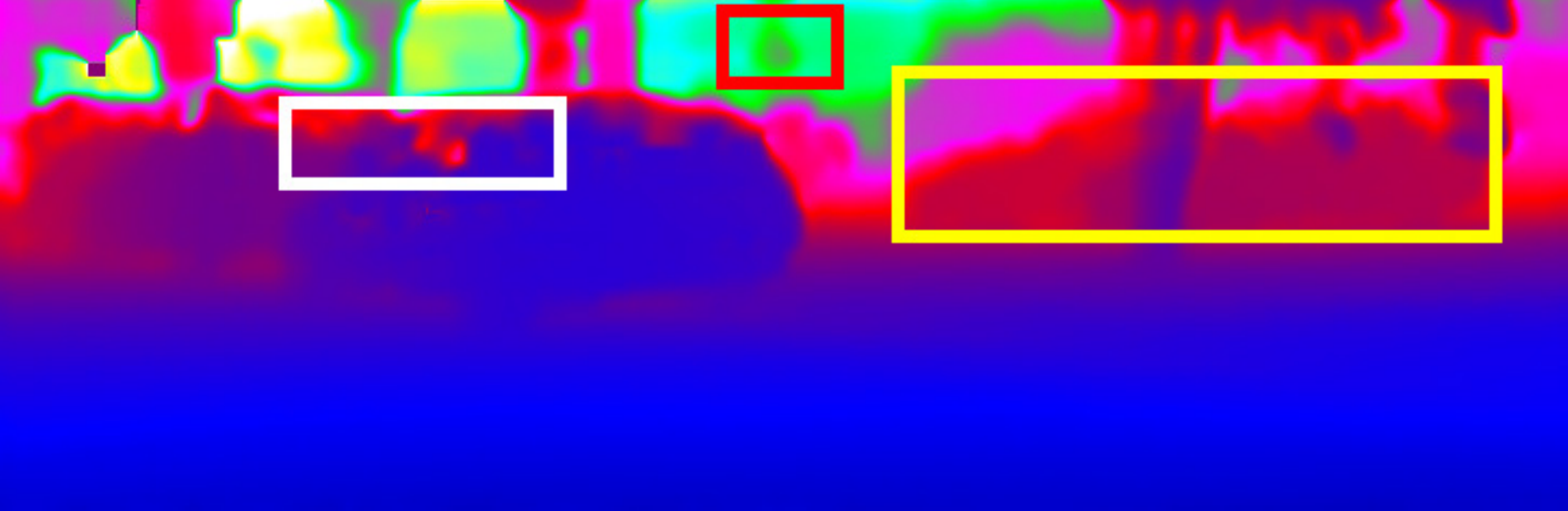} \\ 
(c) Result by ADNN \cite{chodosh2018deep} & (d) Result by SPConv \cite{Uhrig2017THREEDV} \\ 
\includegraphics[height=2.25cm]{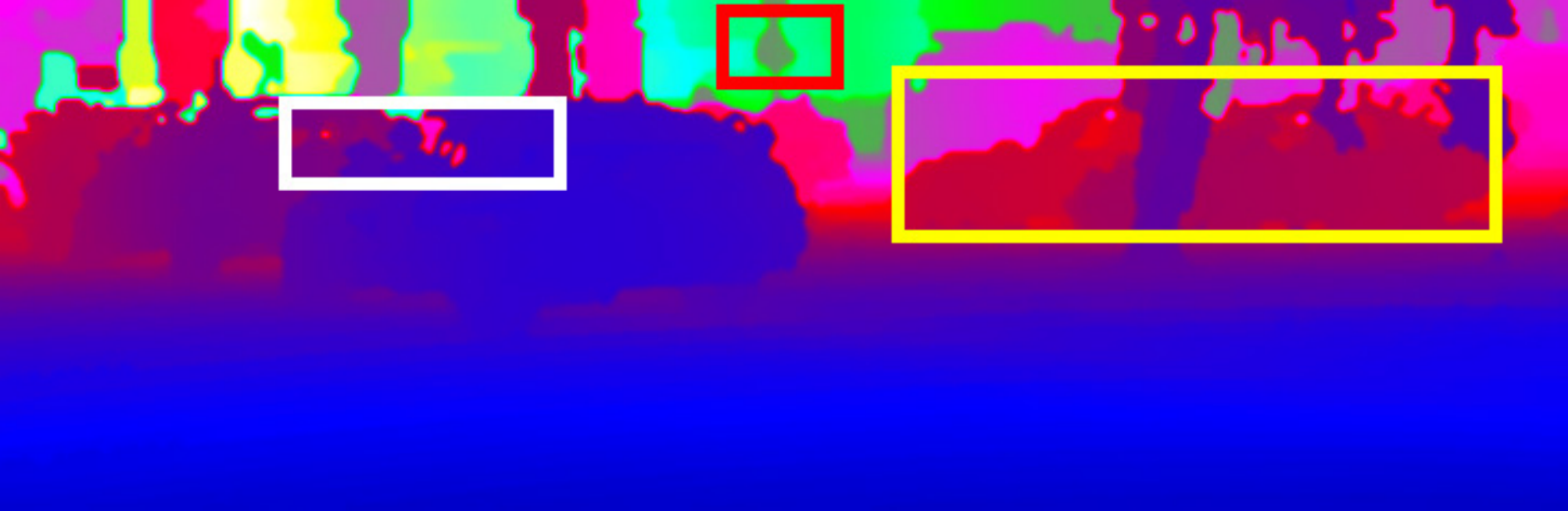} & \includegraphics[height=2.25cm]{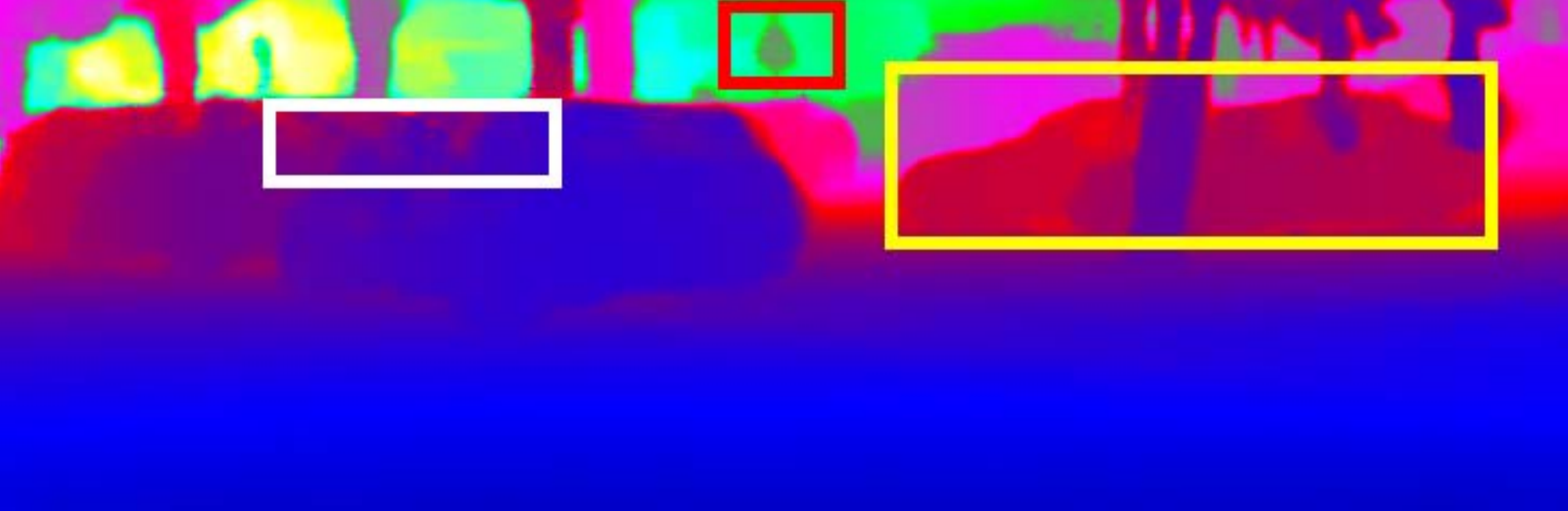} \\ 
(e) Result by IP-Basic \cite{ku2018defense} & (f) Result by Proposed HMS-Net
\end{tabular}
\caption{Illustration of depth completion results by previous methods and our proposed HMS-Net. (a) An example input sparse depth map. (b) Corresponding RGB image. (c) Result by ADNN \cite{chodosh2018deep}. (d) Result by sparse convolution \cite{Uhrig2017THREEDV}. (e) Result by hand-crafted classical image processing method \cite{ku2018defense}. (f) Result by the proposed HMS-Net.}
\label{fig:intro}
\end{figure*}

For learning-based approaches, state-of-the-art methods are mainly based on deep neural networks. Previous methods mainly utilized deep convolutional neural networks (CNN) for generating dense depth maps from sparse inputs. Ma and Karaman \cite{ma2017sparse} simply filled $0$s to locations without depth inputs to create dense input maps, which might introduce ambiguity to very small depth values. Chodosh et al. \cite{chodosh2018deep} proposed to extract multi-level sparse codes from the inputs and used a 3-layer CNN for depth completion. However, those two methods used conventional convolution operations designed for dense inputs (see Fig. \ref{fig:intro}(c) for an example). Uhrig et al. \cite{Uhrig2017THREEDV} proposed sparsity-invariant convolution, which is specifically designed to process sparse maps and enables processing sparse inputs more effectively with CNN.

However, the sparsity-invariant convolution in \cite{Uhrig2017THREEDV} only mimics the behavior of convolution operations in conventional dense CNNs. Its feature maps of later stages lose much spatial information and therefore cannot effectively integrate both low-level and high-level features for accurate depth completion (see Fig. \ref{fig:convnet_encoder_decoder}(a) for illustration). On the other hand, there exist effective multi-scale encoder-decoder network structures for dense pixel-wise classification tasks (see Fig. \ref{fig:convnet_encoder_decoder}(b)), such as U-Net \cite{ronneberger2015u}, Feature Pyramid Network \cite{Lin2017CVPR}, Full Resolution Residual Network \cite{pohlen2017full}. Direct integration of the sparsity-invariant convolution in \cite{Uhrig2017THREEDV} into the multi-scale structures is infeasible, as those structures also require other operations for multi-scale feature fusion, such as sparsity-invariant feature upsampling, average, and concatenation.

To overcome such limitation, we propose three novel sparsity-invariant operations to enable using encoder-decoder networks for depth completion. The three novel operations include sparsity-invariant upsampling, sparsity-invariant average, and joint sparsity-invariant concatenation and convolution. To effectively and efficiently handle sparse feature maps, sparsity masks are utilized at all locations of the feature maps. They record the locations of the sparse features at the output of each processing stage and guide the calculation of the forward and backward propagation. Each sparsity-invariant operation is designed to properly maintain and modify the sparisity masks across the network. The design of those operations are non-trivial and are the keys to using encoder-decoder structures with sparse features. Based on such operations, we propose a multi-scale encoder-decoder network, HMS-Net, which adopts a series of sparsity-invariant convlutions with downsampling and upsampling to generate multi-scale feature maps and shortcut paths for effectively fusing multi-scale features. Extensive experiments on KITTI \cite{Uhrig2017THREEDV} and NYU-depth-v2 \cite{silberman2012indoor} datasets show that our algorithm achieves state-of-the-art depth completion accuracy. 

The main contributions of our work can be summarized to threefold.
1) We design three sparsity-invariant operations for handling sparse inputs and feature maps, which are important to handle sparse feature maps. 2) Based on the proposed sparsity-invariant operations, a novel hierarchical multi-scale network structure fusing information from different scales is designed to solve the depth completion task. 3) Our method outperforms state-of-the-art methods in depth completion. On KITTI depth completion benchmark, our method without RGB information ranks \emph{1st} among all peer-reviewed methods with RGB inputs, while our method with RGB guidance ranks \emph{2nd} among all RGB-guided methods.

\section{Related work}

\subsection{Depth completion}

Depth completion is an active research area with a large number of applications. According to the sparsity of the inputs, current methods could be divided into two categories: sparse depth completion and depth enhancement. The former methods aim at recovering dense depth maps from spatially sparse inputs, while the later methods work on conventional RGB-D depth data and focus on filling irregular and relatively small holes in input dense depth maps. Besides, if the input depth maps are regularly sampled, the depth completion task could be regarded as a depth upsampling (also known as depth super-resolution) task. In other words, depth upsampling algorithms handle a special subset of the depth completion task. The inputs of depth upsampling are depth maps of lower resolution. According to whether RGB information is utilized, depth upsampling methods could be divided into two categories: guided depth upsampling and depth upsampling without guidance.

\subsubsection{Sparse depth completion}

This type of methods take sparse depth maps as inputs.
To handle sparse inputs and sparse intermediate feature maps, Uhrig et al. \cite{Uhrig2017THREEDV} proposed sparsity-invariant convolution to replace the conventional convolution in convolution neural networks (CNN). The converted sparsity-invariant CNN keeps track of sparsity masks at each layer and is able to estimate dense depth maps from sparse inputs.
Ku et al. \cite{ku2018defense} proposed to use a series of hand-crafted image processing algorithms to transform the sparse depth inputs into dense depth maps. The proposed framework first utilized conventional morphological operations, such as dilation and closure, to make the input depth maps denser. It then filled holes in the intermediate denser depth maps to obtain the final outputs. Eldesokey  et al. \cite{eldesokey2018propagating} proposed an algebraically-constrained normalized convolution operation for handling sparse data and propagate depth confidence across layers. The regression loss and depth confidence are jointly optimized. Ren et al. \cite{ren2018sbnet} focused on another task, the efficiency of convolution, but also have the design of sparsity masks. However, their algorithm could not be used in the scenario where the sparsity of the input varies in a large range. The above mentioned methods did not consider image information captured by the calibrated RGB cameras.

There also exist works utilizing RGB images as additional information to achieve better depth completion.
Schneider et al. \cite{schneider2016semantically} combined both intensity cues and object boundary cues to complete sparse depth maps. 
Liao et al. \cite{liao2017parse} utilized a residual neural network and combine the classification and regression losses for depth estimation with both RGB and sparse depth maps as inputs. 
Ma and Karaman \cite{ma2017sparse} proposed to use a single deep regression network to learn directly from the RGB-D raw data, where the depth channel only has sparse depth values. However, the proposed algorithm mainly focused on the indoor scenes and was only tested on the indoor NYUv2 dataset.  Instead, Van Gansbeke et al. \cite{van2019sparse} combine RGB and depth information through summation according to two predicted confidence maps.
Chodosh et al. \cite{chodosh2018deep} utilized compressed sensing techniques and Alternating Direction Neural Networks to create a deep recurrent auto-encoder for depth completion. Sparse codes were extracted at the outputs of multiple levels of the CNN and were used to generate dense depth prediction. 
Zhang and Funkhouser \cite{zhang2018deep} adopted a neural network to predict dense surface normals and occlusion boundaries from RGB images. These predictions were then used as auxiliary information to accomplish the depth completion task from sparse depth data. 
Qiu et al. \cite{qiu2019deeplidar} further extended this idea on outdoor datasets by generating surface normals as intermediate representation.
Jaritz et al. \cite{Jaritz20183DV} argued that by using networks with large receptive field, the networks were not required to have special treatment for sparse data. They instead trained networks with depth maps of different sparsities. 
Cheng et al. \cite{Cheng2018ECCV} proposed to learn robust affinities between pixels to spatially propagate depth values via convolution operations to fulfill the entire depth map, 
while Eldesokey el al. \cite{eldesokey2019confidence} used continuous confidence instead of validity maps and algebraically constrained filters to tackle the sparsity-invariance problem and also guidance from both RGB images and the output confidence produced by their unguided network.
Yang et al. \cite{yang2019dense} proposed to yield the full posterior over depth maps with a Conditional Prior Network from their previous work \cite{yang2018conditional}.
And a self-supervised training framework was designed by Ma et al. \cite{Ma2018arxiv}, which explores temporal relations of video sequences to provide additional photometric supervisions for depth completion networks.

\subsubsection{Depth enhancement}

The inputs of depth enhancement or depth hole-filling methods are usually dense depth maps with irregular and rare small holes. The input depth maps are usually captured with RGB images. 
Matyunin et al. \cite{matyunin2011temporal} used the depth from the neighborhoods of the hole regions to fill the holes, according to the similarity of RGB pixels. In \cite{camplani2012efficient}, the missing depth values are obtained by iteratively applying a joint-bilateral filter to the hole regions' neighboring pixels. Yang et al. \cite{Yang:253660} propose an efficient depth image recovery algorithm based on auto-regressive correlations and recover high-quality multi-view depth frames with it. They also utilized color image, depth maps from neighboring views and temporal adjacency to help the recovery. Chen et al.\cite{chen2012depth} firstly created smooth regions around the pixels without depth, and then adopted a bilateral filter without smooth region constraints to fill the depth values. Yang et al. \cite{yang2014color} proposed an adaptive color-guided autoregressive model for depth enhancement, which utilized local correlation in the initial depth map and non-local similarity in the corresponding RGB image.

\subsubsection{Guided depth upsampling}

Depth upsampling methods take low-resolution depth maps as inputs and output high-resolution ones.
As the guidance signals, the provided RGB images bring valuable information (e.g., edges) for upsampling. Li et al. \cite{li2016deep} proposed a CNN to extract features from the low-resolution depth map and the guidance image to merge their information for estimating the upsampled depth map. In \cite{ferstl2013image}, an anisotropic diffusion tensor is calculated from a high-resolution intensity image to serve as the guidance. An energy function is designed and solved to solve the depth upsampling problem. Hui et al. \cite{hui2016depth} proposed a convolution neural network, which fused the RGB guidance signals at different stages. They trained the neural network in the high-frequency domain. Xie et al. \cite{xie2016edge} upsampled the low-resolution depth maps with the guidance of image edge maps and a Markov Random Field model. Jiang et al. \cite{guo2018hierarchical} proposed a depth super-resolution algorithm utilizing transform domain regularization with an auto-regressive model, as well as a spatial domain regularization by injecting a multi-directional total variation prior.

\subsubsection{Depth upsampling without guidance}

Depth upsampling could also be achieved without the assistance of corresponding RGB images. As an MRF based method, the approach in \cite{mac2012patch} matched against the height field of each input low-resolution depth patch, and searched the database for a list of most appropriate high-resolution candidate patches. Selecting the correct candidates was then posed as a Markov Random Field labeling problem. Ferstl et al. \cite{ferstl2015variational} learned a dictionary of edge priors from an external database of high- and low-resolution depth samples, and utilized a novel variational sparse coding approach for upsampling. Xie et al. \cite{xie2015joint} used a coupled dictionary learning method with locality coordinate constraints to transform the original depth maps into high-resolution depth maps. Riegler et al. \cite{riegler2016atgv} integrated a variational method that modeled the piecewise affine structures in depth maps on top of a deep network for depth upsampling.

\subsection{Multi-scale networks for pixelwise prediction}

Neural networks that utilize multi-scale feature maps for pixelwise prediction (e.g., semantic segmentation) were widely investigated. Combining both low-level and high-level features was proven to be crucial for making accurate pixelwise prediction. 

Ronneberger et al. \cite{ronneberger2015u} proposed a U-shaped network (U-Net). It consists of an iterative downsampling image encoder that gradually summarized image information into smaller but deeper feature maps, and an iterative upsampling decoder that gradually combined low-level and high-level features to output the pixelwise prediction maps. The low-level information from the encoder was passed to the high-level information of the decoder by direct concatenation of the feature maps of the same spatial sizes. Such a network has shown its great usefulness in many 2D and 3D segmentation tasks. Similar network structures, which include Hourglass \cite{Newell2016ECCV} and Feature Pyramid Network \cite{Lin2017CVPR}, have also been investigated to tackle pixelwise prediction tasks. Recently, Pohlen et al. \cite{pohlen2017full} proposed the full-resolution residual network (FRRN), which treated full-resolution information as a residual flow to pass valuable information across different scales for semantic segmentation. He et al. \cite{he2018learning} designed a fully fused network to utilize both images and focal length to learn the depth map. 

However, even with the sparsity-invariant convolution operation proposed in \cite{Uhrig2017THREEDV}, the multi-scale encoder-decoder networks cannot be directly converted to handle sparse inputs. This is because there exist many operations that do not support sparse feature maps. Our proposed sparsity-invariant operations solve the problem and allow encoder-decoder networks to be used for sparse data.

\section{Method}
\label{sec:method}

We introduce our proposed framework for depth completion in this section. In Section \ref{ssec:si_convolution}, we first review sparsity-invariant convolution proposed in \cite{Uhrig2017THREEDV}. In Section \ref{ssec:si_operations}, we then introduce three novel sparsity-invariant operations, which are crucial for adopting multi-scale encoder-decoder networks to process sparse inputs. In Section \ref{ssec:hms_net}, based on such sparsity-invariant operators, the hierarchical multi-scale encoder-decoder network (HMS-Net) for effectively combining multi-scale features is proposed to tackle the depth completion task.

\subsection{Sparsity-invariant Convolution}
\label{ssec:si_convolution}

In this subsection, we first review sparsity-invariant convolution in \cite{Uhrig2017THREEDV}, which modifies conventional convolution to handle sparse input feature maps. The sparsity-invariant convolution is formulated as
\begin{align}
\boldsymbol{z}(u,v)=\frac{\sum_{i,j=-k}^k \boldsymbol{m_x}(u+i,v+j) \boldsymbol{w}(i,j) \boldsymbol{x}(u+i,v+j)}{\sum_{i,j=-k}^k \boldsymbol{m_x}(u+i,v+j)+\epsilon}+b.
\label{eqn:si_conv}
\end{align}
The sparisity-invariant convolution takes a sparse feature map $\boldsymbol{x}$ and a binary single-channel sparsity mask $\boldsymbol{m}$ as inputs, which both has the same spatial size $H \times W$. The convolution generates output features $\boldsymbol{z}(u,v)$ for each location $(u,v)$. At each spatial location $(u,v)$, the binary sparsity mask $\boldsymbol{m_x}(u,v)$ records whether there are input features at this location, i.e., $1$ for features existence and $0$ otherwise. The convolution kernel $\boldsymbol{w}$ is of size $(2k+1) \times (2k+1)$, and $b$ represents a learnable bias vector. Note that the kernel weights $\boldsymbol{w}$ and bias vector $b$ are learned via back-propagation, while the sparsity mask $\boldsymbol{m_x}$ is specified by the previous layer and is trained. 

The key difference with conventional convolution is the use of binary sparsity mask $\boldsymbol{m_x}$ for convolution calculation. The mask values on numerator of Eq. \eqref{eqn:si_conv} denote that when conducting convolution, only input features at the valid or visible locations specified by the sparsity mask $\boldsymbol{m_x}$ are considered. The mask values in denominator denote that, since only a subset of input features are involved, the output features should be normalized according to the number of valid input locations. $\epsilon$ represents a very small number and is used to avoid division by 0.

Note that the sparsity mask should always indicate the validity or sparsity of each location of the feature maps. Since the convolution layers in a neural network is generally stacked for multiple times, the output sparsity mask $\boldsymbol{m_z}$ at each stage should be modified to match the output of each stage $\boldsymbol{z}$. For each output feature location $(u,v)$, if there is at least one valid input location in its receptive field of the previous input, its sparsity mask $\boldsymbol{m_z}(u,v)$ should be updated to $1$. In practice, the output sparsity mask is obtained by conducting max pooling on the input sparsity mask with the same kernel size of convolution $(2k+1) \times (2k+1)$.

\begin{figure}
\centering
\includegraphics[height=3.5cm]{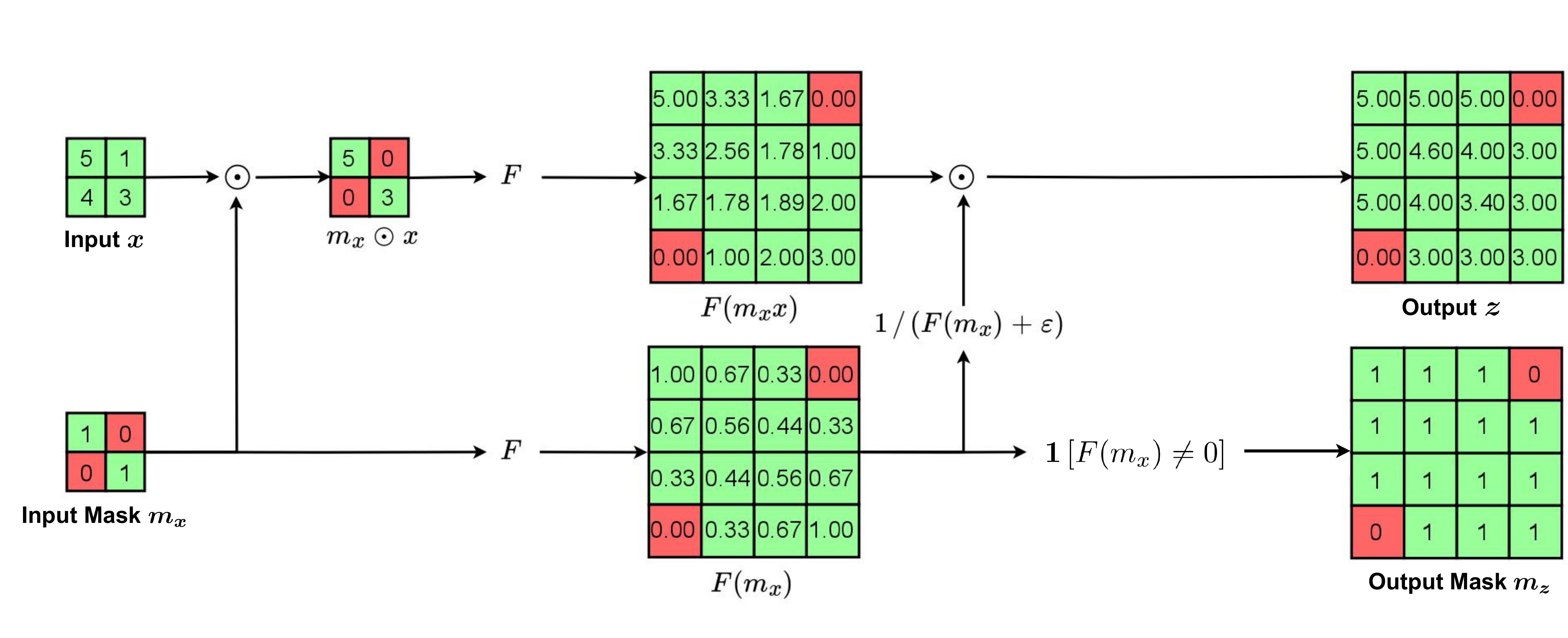}
\caption{Illustration of the proposed sparsity-invariant upsampling operation. $\boldsymbol{F}$ stands for Bilinear Upsampling.}
\label{fig:si_upsampling}
\end{figure}

\subsection{Sparsity-invariant operations}
\label{ssec:si_operations}

The sparsity-invariant convolution successfully converts conventional convolution to handle sparse input features and is able to stack multiple stages for learning highly non-linear functions. However, only modifying convolution operations is not enough if one tries to utilize state-of-the-art multi-scale encoder-decoder structure for pixelwise prediction. As shown in Fig. \ref{fig:convnet_encoder_decoder}(b), average, upsampling, and concatenation are also common operations in a multi-scale encoder-decoder networks. Therefore, we propose the sparsity-invariant version of these operations: sparsity-invariant average, sparsity-invariant upsampling, and joint sparsity-invariant concatenation and convolution. The three sparsity-invariant operations allow effectively handling sparse feature maps across the entire encoder-decoder network. They are the foundations of complex building blocks in our overall framework.

Designing the sparsity-invariant operations is non-trivial. Our proposed sparsity-invariant operations share the similar spirit of sparsity-invariant convolution: using single-channel sparsity masks to track the validity of feature map locations. The sparsity masks could be used to guide and regularize the calculation of the operations.

\subsubsection{Sparsity-invariant bilinear upsampling}
\label{sssec:si_upsample}

One of the most important operations in encoder-decoder networks is the upsampling operation in the decoder part. We first propose the sparsity-invariant bilinear upsampling operation. Let $\boldsymbol{x}$ and $\boldsymbol{m_x}$ denote the input sparse feature map and the corresponding input sparsity mask of size $H \times W$. The operation generates the output feature map $\boldsymbol{z}$ and its corresponding sparsity mask $\boldsymbol{m_z}$ of size $2H \times 2W$. Let $F$ represents the conventional bilinear upsampling operator, which bilinearly upsamples the input feature map or mask by two times. The proposed sparsity-invariant bilinear upsampling can be formulated as 
\begin{align}
\boldsymbol{z} &= \frac{F(\boldsymbol{m_x} \odot \boldsymbol{x})}{F (\boldsymbol{m_x})+\epsilon}, \label{eq:si_upsample_output} \\
\boldsymbol{m_z}&= \boldsymbol{1} \left[ F(\boldsymbol{m_x}) \neq 0 \right], \label{eq:si_upsample_mask}
\end{align}
where $\odot$ denotes the spatial elementwise multiplication, $\epsilon$ is a very small number to avoid division by zero, and $\boldsymbol{1} [\cdot]$ denotes the indicator function, i.e., $\boldsymbol{1} [\text{true}] = 1$ and $\boldsymbol{1} [\text{false}] = 0$. The proposed sparsity-invariant bilinear upsampling operation is illustrated in Fig. \ref{fig:si_upsampling}.

As shown by Eq. \eqref{eq:si_upsample_output}, the proposed operation first uses the input sparsity mask $\boldsymbol{m_x}$ to mask out the invalid features from the input feature maps $\boldsymbol{x}$ as $\boldsymbol{m_x} \odot \boldsymbol{x}$. The traditional bilinear upsampling $F$ operator is then applied to upsample both the masked feature maps $\boldsymbol{m_x} \odot \boldsymbol{x}$ and the input sparsity mask $\boldsymbol{m_x}$. The upsampled sparse features $F(\boldsymbol{m_x} \odot \boldsymbol{x})$ are then normalized at each location according to the upsampled sparsity mask values $F(\boldsymbol{m_x})$. The final sparsity mask $\boldsymbol{m_z}$ is obtained by identifying the non-zero locations of the upsampled sparsity mask $F(\boldsymbol{m_x})$.

Note that for sparsity-invariant max-pooling or downsampling, it could be calculated the same as Eqs. \eqref{eq:si_upsample_output} and \eqref{eq:si_upsample_mask} by replacing the upsampling function $F$ with max-pooling or downsampling operators.


\begin{figure}
\centering
\includegraphics[height=4.5cm]{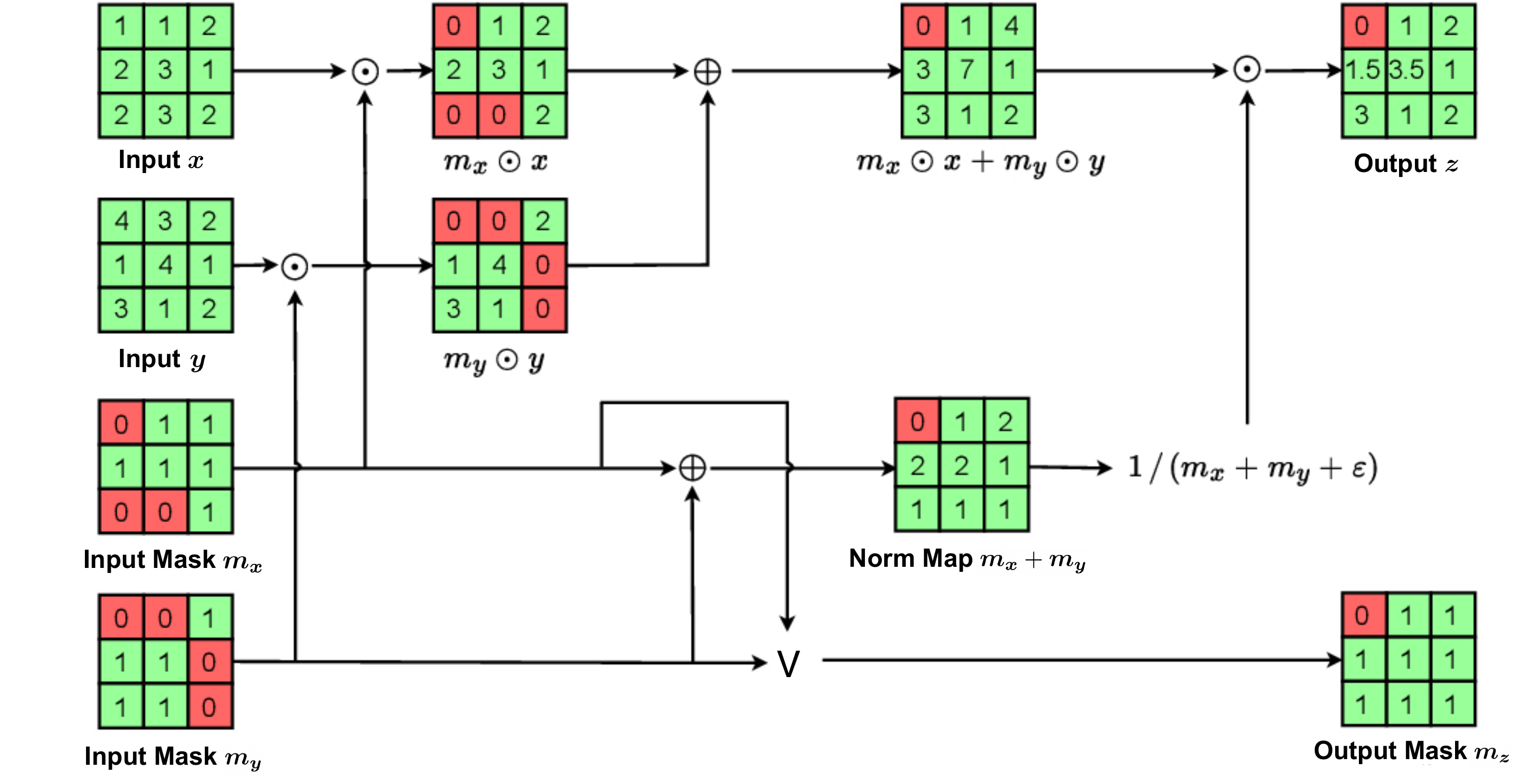}
\caption{Illustration of the proposed sparsity-invariant average.}
\label{fig:si_addition}
\end{figure}

\subsubsection{Sparsity-invariant average}

Pixelwise average of two feature maps of the same spatial sizes is needed for fusing features from different levels without increasing the output channels. For average of sparse input feature maps, however, specifically designed average operation is needed. We propose sparsity-invariant average, which takes two input sparse feature maps, $\boldsymbol{x}$ and $\boldsymbol{y}$, with their corresponding sparsity masks, $\boldsymbol{m_x}$ and $\boldsymbol{m_y}$, as inputs. It generates the fused sparse feature maps $\boldsymbol{z}$ with its corresponding sparsity mask $\boldsymbol{m_z}$. Unlike the sparsity-invariant upsampling or downsampling, the key difference is that the average operation takes two sparse features as inputs.

We formulate the sparsity-invariant average as
\begin{align}
&\boldsymbol{z}= \frac{\boldsymbol{m_x} \odot {\boldsymbol{x}} + \boldsymbol{m_y} \odot \boldsymbol{y}}{\boldsymbol{m_x} + \boldsymbol{m_y} + \epsilon}, \label{eq:si_addition_output}\\
&\boldsymbol{m_z}=\boldsymbol{m_x}\vee\boldsymbol{m_y},
\label{eq:si_addition_mask}
\end{align}
where $\vee$ denotes elementwise alternation (i.e., logical ``or'') function, $\odot$ represents elementwise multiplication, and $\epsilon$ is a very small number to avoid division by zero.

The sparsity-invariant average is illustrated in Fig. \ref{fig:si_addition}. The two input sparse features are first masked out by their corresponding masks to obtain $\boldsymbol{m_x} \odot {\boldsymbol{x}}$ and $\boldsymbol{m_y} \odot \boldsymbol{y}$. Both the masked features and the masks are then pixelwisely added. The added features are then normalized by added sparsity masks to calculate the output sparse features $\boldsymbol{z}$. For the output sparsity mask, at each location, if the location is valid for either of the input feature maps, the mask is set to 1 for this location. At each location of the output feature map, the output feature vector is the mean of two input feature vectors if both input maps are valid at this location. If only one input feature is valid at a location, the valid single input feature vector is directly copied to the same location of the output feature maps. The two types of output feature vectors would have similar magnitude because the added features are properly normalized by the added sparsity mask. 

Note feature addition was also explored in \cite{Uhrig2017THREEDV}. However, an output sparsity mask was not generated. Without such a mask, following convolution operations cannot handle sparse output features. Therefore, their operation could only be used as the last layer of a neural network.

\subsubsection{Joint sparsity-invariant concatenation and convolution}
\label{sssec:si_concat}
\begin{figure}
	\centering
	\includegraphics[height=4.0cm]{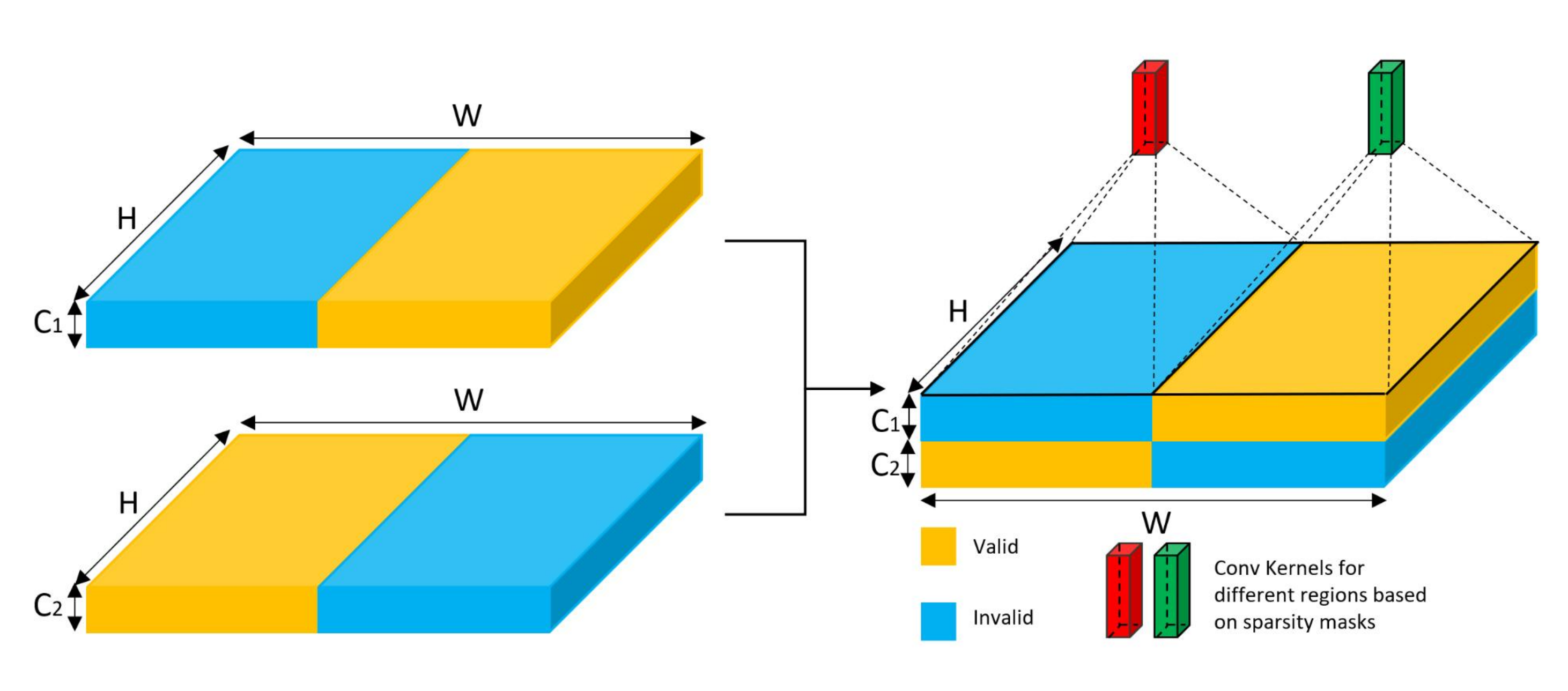}
	\caption{Sparsity patterns vary from regions, thus we need several different kernels to deal with out feature maps after concatenation.}
	\label{fig:si_concat_problem}
\end{figure}

\begin{figure}
	\centering
	\includegraphics[height=4.8cm]{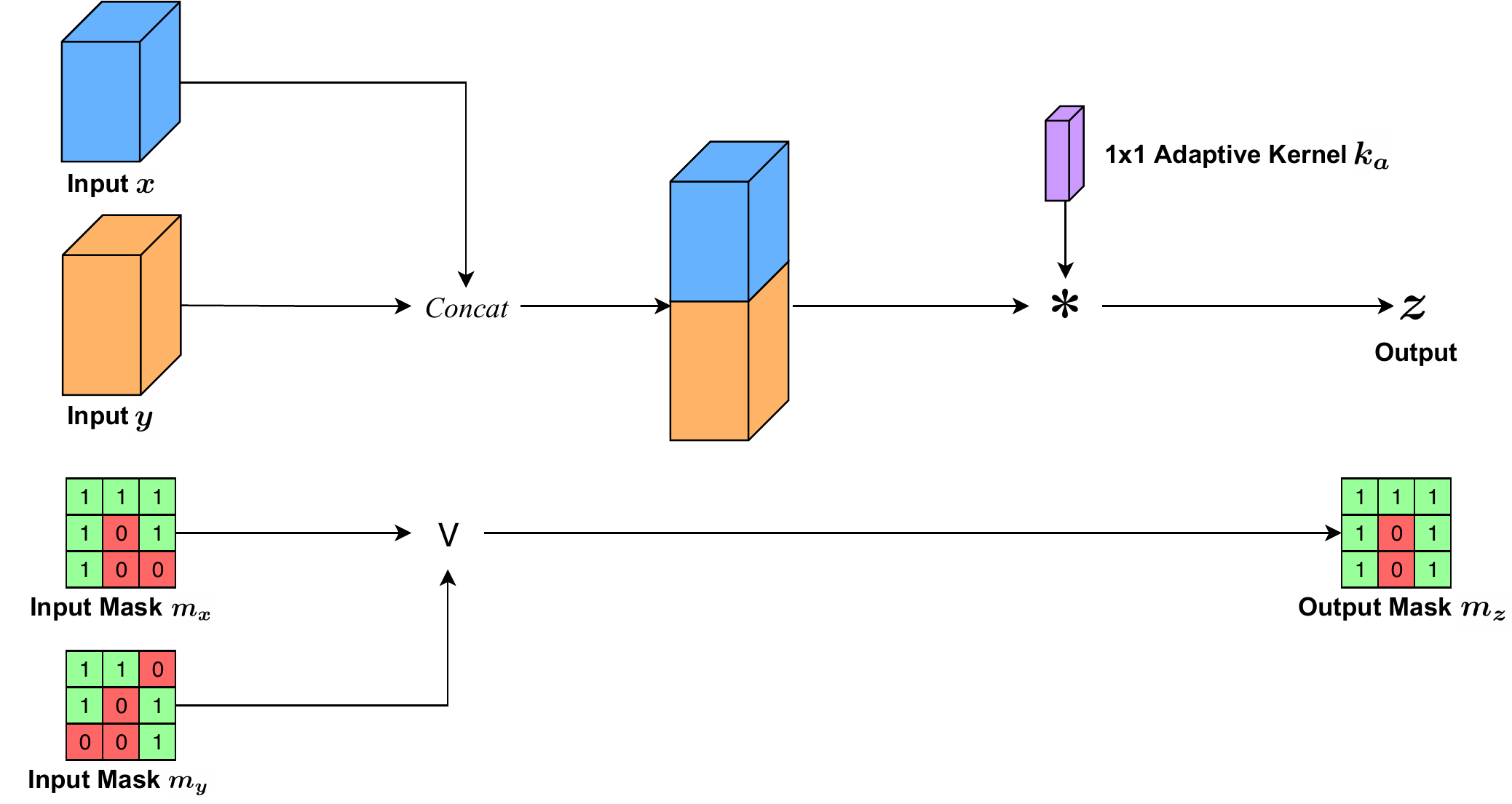}
	\caption{Illustration of the proposed joint sparsity-invariant concatenation and convolution.}
	\label{fig:si_concat}
\end{figure}
Another commonly used approach of fusing two feature maps with the same spatial size is feature concatenation, i.e., concatenation along the channel dimension. 
However, different from aforementioned other operations, concatenation would introduce sparsity in both the spatial dimension (H $\times$ W) and the feature dimension (C), and the latter actually prevent us from simply extending the idea of sparsity-invariant average in the previous subsection. Let's consider the scenario where two feature maps with shape $C_1 \times H \times W$ and $C_2 \times H \times W$ are now being concatenated into one with shape $(C_1+C_2) \times H \times W$ as illustrated in Fig. \ref{fig:si_concat_problem}. The concatenation is further followed by a convolution layer ($1\times 1$ for simplicity), which is a common operation in CNNs. However, we know that convolution performs filtering on a location by extracting one local feature vector with length $C$ and summing all entries up into a number with learnable weights. Then, the convolution kernels iterate over the whole feature map, treating every location equally. 

Recall the situation in sparsity-invariant convolution where the feature vector for a certain location only have two possible sparsity patterns-the whole vector of length $C$ is valid, or all the entries of this vector are zeros. Note that the latter situation would not influence the training as its contribution to the output as well as the gradient of kernels is zero. Therefore, it is enough for us to use one set of convolution kernels, equally for every valid location. 

However, when we are convolving on the feature maps after concatenation, we have four different types of vectors or sparsity patterns for each location: the first $C_1$ feature channels of the vector is valid while the latter $C_2$ feature channels are not; or $C_2$ is valid while $C_1$ is not, or both of them are valid/invalid. Therefore, we need three different sets of kernels to tackle these four different sparsity patterns. In other words, to effectively handle different scenarios at different locations of the concatenated feature maps, we propose to use an adaptive-kernel version convolution to solve the difficulty and combine it with concatenation together. 

Another advantage of combining them is that all convolution kernels would generate outputs with the same spatial sparsity patterns. Therefore the output mask is still of single channel, which is computationally efficient and reduces the model complexity significantly. Specifically, our joint sparsity-invariant concatenation and convolution is described and explained formally as following: 

\begin{figure*}[t]
\centering
\footnotesize
\begin{tabular}{c@{\hspace{-4.5mm}}c@{\hspace{4mm}}c}
	&\includegraphics[scale=0.36]{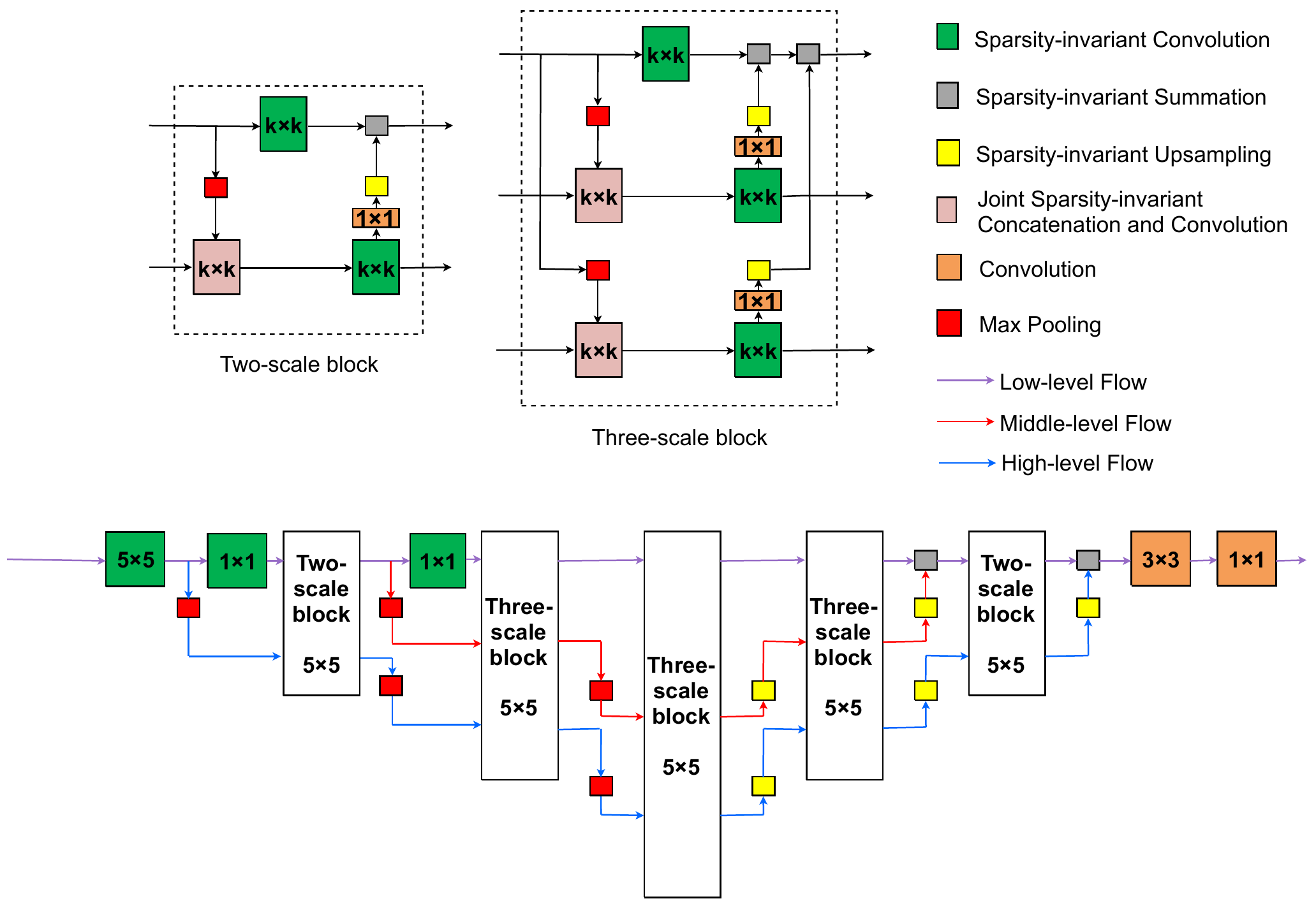}
	& \includegraphics[scale=0.36]{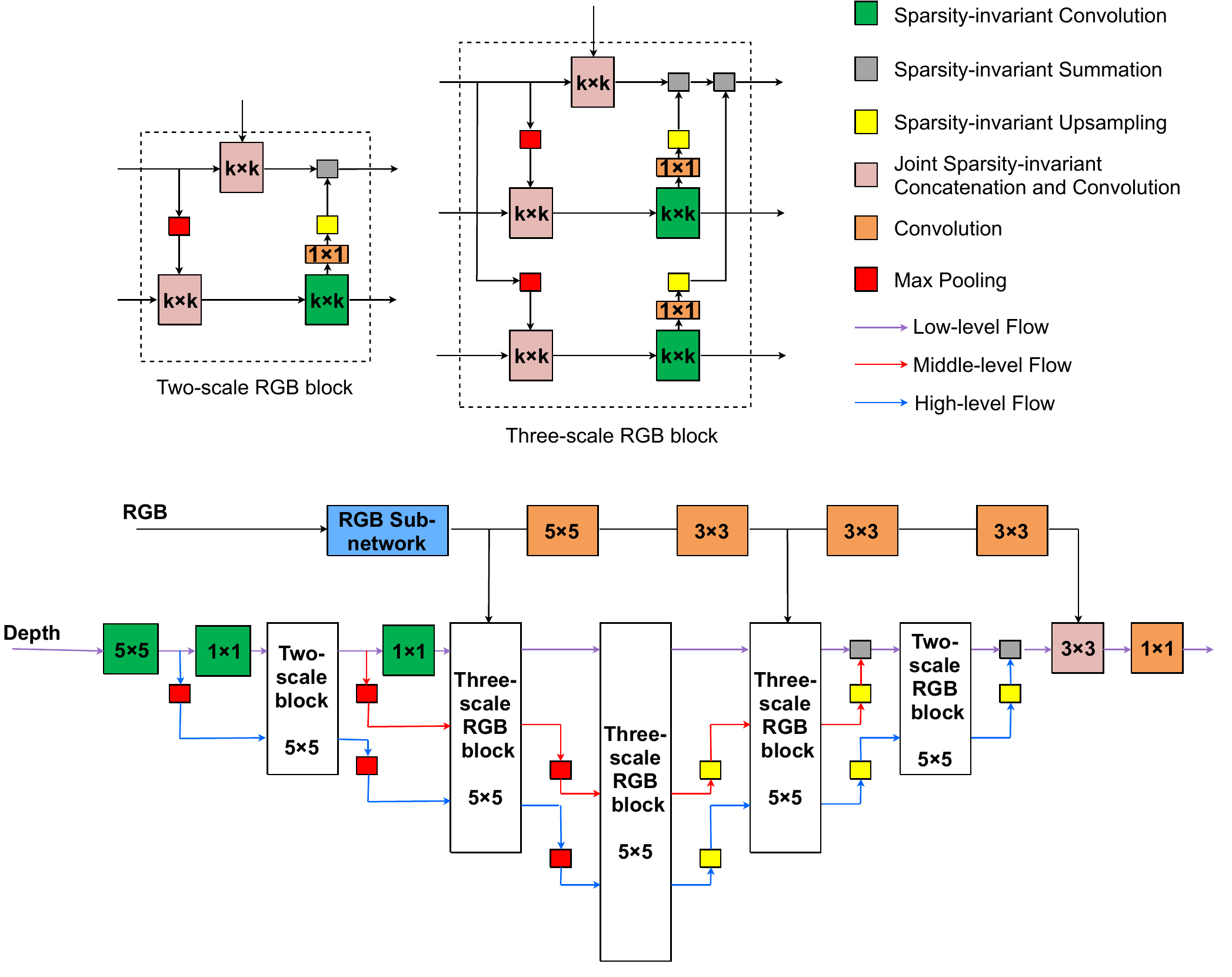}\\
	& (a) Sparsity-invariant network without RGB guidance & (b) Sparsity-invariant network with RGB guidance
\end{tabular}
\caption{Illustration of our multi-scale encoder-decoder network structure for depth completion based on the proposed sparsity-invariant operations. (a) Proposed network with RGB guidance. (b) Proposed network without RGB guidance.}
\label{fig:overall}
\end{figure*}

Given the two input sparse feature maps $\boldsymbol{x}$ and $\boldsymbol{y}$ with their sparsity masks $\boldsymbol{m_x}$ and $\boldsymbol{m_x}$, the proposed joint concatenation and convolution operation is formulated as
\begin{align}
&\boldsymbol{z} = \left[ \boldsymbol{x}; \boldsymbol{y} \right] * \boldsymbol{k_a},\\
&\boldsymbol{m_z}= \boldsymbol{m_x}\vee\boldsymbol{m_y},
\label{eq:si_addition_mask}
\end{align}
where $[;]$ denotes the concatenation of two feature maps along the channel dimension, and $*$ denotes the conventional convolution operation. Note that the output sparsity mask is calculated exactly the same as that in sparsity-invariant average. The key of the proposed operation is a $1\times 1$ convolution with an adaptive convolution kernel $\boldsymbol{k_a}$ that handles three different scenarios of concatenating sparse feature maps, which is formulated as
\begin{align}
\boldsymbol{k_a} (u,v) =
\begin{cases}
\boldsymbol{k}_{\boldsymbol{a}}^{(1)} & \boldsymbol{m_x}(u,v)=1, \, \boldsymbol{m_y}(u,v)=0;\\
\boldsymbol{k}_{\boldsymbol{a}}^{(2)} & \boldsymbol{m_x}(u,v)=0, \, \boldsymbol{m_y}(u,v)=1;\\
\boldsymbol{k}_{\boldsymbol{a}}^{(3)} & \boldsymbol{m_x}(u,v)=1, \, \boldsymbol{m_y}(u,v)=1,
\end{cases}
\end{align}
where $\boldsymbol{k_a} (u,v)$ are the $1\times 1$ adaptive convolution kernel at location $(u,v)$ of the concatenated feature maps $[\boldsymbol{x}; \boldsymbol{y}]$. $\boldsymbol{k}_{\boldsymbol{a}}^{(1)}$ $\boldsymbol{k}_{\boldsymbol{a}}^{(2)}$, $\boldsymbol{k}_{\boldsymbol{a}}^{(3)}$ are the three sets of learnable convolution weights for the three different feature concatenation scenarios: at each location $(u,v)$, either both input feature vectors are valid (i.e., $\boldsymbol{m_x}(u,v)=1$ and  $\boldsymbol{m_y}(u,v)=1$), or only one of the input feature vectors is valid (i.e., either $\boldsymbol{m_x}(u,v)=1$ or $\boldsymbol{m_y}(u,v)=1$). 
The key reason for using different sets of kernel weights instead of the same set of convolution weights, as we briefly introduced before, is to avoid involving invalid input features in the concatenated feature maps into feature learning process. 
Illustrating with our notation above, if the current 1 $\times$ 1 convolution kernel is on the location $(u, v)$ and find that the first mask here $\boldsymbol{m_x}(u,v)$ is one and the second mask $\boldsymbol{m_y}(u,v)$ is zero, it would choose the first set of kernel weights $\boldsymbol{k}_{\boldsymbol{a}}^{(1)}$ to use during forward pass. So is the back propagation. In this case, we know the second chunk of the feature vector, of which the length is fixed, is always zero. And because it's consistently processed by the first kernel, this kernel would naturally learn how to adapt to this pattern.

In other words, by adopting the proposed adaptive convolution kernel $\boldsymbol{k_a}$, the three sets of kernel weights $\boldsymbol{k}_{\boldsymbol{a}}^{(1)}$, $\boldsymbol{k}_{\boldsymbol{a}}^{(2)}$, $\boldsymbol{k}_{\boldsymbol{a}}^{(3)}$ are able to handle different sparse feature concatenation scenarios. With joint training, the different kernels are learned to best adapt each other to generate appropriate feature representations for further processing. In this way, the sparse feature maps could be effectively fused with proposed concatenation.

\subsection{Hierarchical Multi-scale Network (HMS-Net) for depth completion}
\label{ssec:hms_net}

Multi-scale encoder-decoder neural networks for dense inputs are widely investigated for pixelwise prediction tasks. Those networks have the advantages of fusing both low-level and high-level features for accurate pixel prediction. However, with only the sparsity-invariant convolution in \cite{Uhrig2017THREEDV}, encoder-decoder networks cannot be converted to handle sparse inputs. On the other hand, for frequently studied pixelwise prediction tasks, such as semantic segmentation, global high-level information usually shows greater importance to the final performance, while the full-resolution low-level features are less informative and generally go through fewer non-linearity layers compared with high-level features. However, we argue that depth completion is a low-level vision task. The low-level features in this task should be non-linearly transformed and fused with mid-level and high-level features for more times to achieve satisfactory depth completion accuracy.

Based on this motivation, we propose the Hierarchical Multi-scale encoder-decoder Network (HMS-Net) with our proposed sparsity-invariant operations for sparse depth completion. The network structure without RGB information is illustrated in Fig. \ref{fig:overall}(a). We propose two basic building blocks, a two-scale block and a three-scale block, consisting proposed sparsity-invariant operations. The two-scale block has an upper path that non-linearly transforms the full-resolution low-level features by a $k\times k$ sparsity-invariant convolution. The lower path takes downsampled low-level features as inputs for learning higher-level features with another $k\times k$ convolution. We empirically set $k$=5 in all our experiments according to our hyperparameter study. The resulting higher-level features are then upsampled and added back to the full-resolution low-level features.
Compared with the two-scale block, the three-scale block fuses features from two higher levels into the upper low-level feature path to utilize more auxiliary global information. In this way, the full-resolution low-level features are effectively fused with higher-level information and are non-linearly transformed multiple times to learn more complex prediction functions. All feature maps in our network are of 16 channels regardless of scales.

Our final network utilizes a $5 \times 5$ sparsity-invariant convolution at the first layer. The resulting features then go through three of the proposed multi-scale blocks followed by sparsity-invariant max-pooling, and are then upsampled three times to generate the full-resolution feature maps. The final feature maps are then transformed by one $1\times 1$ convolution layers to generate the final per-pixel prediction. The output depth predictions are supervised by the Mean-square Error (MSE) loss function with ground-truth annotations.

Another way to understand the network structure we proposed is that our structure could be considered as a backbone encoder-decoder CNN shown in Fig. \ref{fig:convnet_encoder_decoder}(b) but with special design for the depth completion task. By looking at the details in Fig. \ref{fig:overall}(a) and compare it with commonly used encoder-decoder networks \cite{ronneberger2015u, Lin2017CVPR, Newell2016ECCV}, it's easy to find that there are several key shortcuts (importance demonstrated later in the ablation study) between different flows uniquely in our network for a better fusion as we described before. Except for this aspect, the low-level features in the mentioned encoder-decoder networks go through very few non-linear transformations, while our proposed network emphasizes much more on the low-level features. Furthermore, the total depth of our network is also much shallower. Compared with Full-Resolution Residual Network \cite{pohlen2017full} which also has multiple shortcuts, the latter's full-resolution low-level features only serve as residual signals. In addition, it does not consider the fusion of multiple-scale features at the same time as our three-scale block does. We further compared the proposed network with the commonly used encoder-decoder network structures in our experimental studies.


\subsection{RGB-guided multi-scale depth completion}

The LIDAR sensors are usually paired with RGB cameras to obtain the aligned sparse depth maps and RGB images. RGB images could therefore act as auxiliary guidance for depth completion. 

To integrate RGB features into our multi-scale encoder-decoder network, we added an RGB feature path to our proposed network. The network structure is illustrated in Fig. \ref{fig:overall}(b). The input image is first processed by an RGB sub-network to obtain mid-level RGB features. The structure of the sub-network follows the first six blocks of the ERFNet \cite{Romera2017IV}. It consists of two downsampling blocks and four residual blocks. The downsampling block has a $2\times 2$ convolution layer with stride 2 and a $2\times 2$ max-pooling layer. The input features are input into the two layers simultaneously, and their results are concatenated along the channel dimension to obtain the $1/2$ size feature maps. The main path of the residual block has two sets of $1\times 3$ conv $\rightarrow$ BN $\rightarrow$ ReLU $\rightarrow$ $3\times 1$ conv $\rightarrow$ BN $\rightarrow$ ReLU. Because the obtained mid-level RGB features are downsampled to $1/4$ of its original size, they are upsamepled back to the input image's original size. The upsampled RGB features are then transformed by a series of convolutions. They act as additional guidance signals and are concatenated to the low-level sparse depth feature maps of different multi-scale blocks. Our experimental results show including the additional RGB mid-level features as guidance further improves the depth completion accuracy.


\subsection{Training scheme}

We adopt the mean squared error (MSE) loss function to train our proposed encoder-decoder networks. Since some datasets could only provide sparse ground-truth depth maps, the loss function is only evaluated at locations with ground-truth annotations, which could be formulated as
\begin{align}
L(\boldsymbol{x}, \boldsymbol{y}) = \frac{1}{|\boldsymbol{V}|} \sum_{u,v\in \boldsymbol{V}} \left| \boldsymbol{o}(u,v) - \boldsymbol{t}(u,v) \right|^2
\label{eq:loss}
\end{align}
where $\boldsymbol{V}$ is the set containing coordinates with ground-truth depth values, $|\boldsymbol{V}|$ calculates the total number of valid points in $\boldsymbol{V}$, and $\boldsymbol{o}$ and $\boldsymbol{t}$ are the predicted and ground-truth depth maps.

For network training, all network parameters except those of the RGB sub-network are randomly initilized. We adopt the ADAM optimizer \cite{kinga2015method} with an initial learning rate of 0.01. The network is trained for 50 epochs. To gradually decrease the learning rate, it is decayed according to the following equation,
\begin{align}
\text{learning rate} = 0.01 \times \left( 1 - \frac{\text{iter\_epoch}}{50} \right)^{0.9},
\end{align}
where iter\_epoch denotes the current epoch iteration. 

Also note that sparsity masks are generated for every input directly depending on the network structure, without any learnable parameters. The values of input masks are set to 1 for all valid spatial locations and 0 for invalid locations. The mask in one layer is propagated to the following layer. In other words, they purely depend on the network structure and the current input. During training, they filter out both invalid feature points and the gradient for invalid spatial locations.

For our RGB sub-network, we use the first six blocks of the ERFNet \cite{Romera2017IV}. Its initial network parameters are copied from the network pretrained on the CityScapes dataset \cite{Cordts2017CVPR}. Both paths are then end-to-end trained until convergence.

\section{Experiments}

We conduct experiments on the KITTI depth completion dataset \cite{Uhrig2017THREEDV} and NYU-depth-v2 dataset \cite{silberman2012indoor} for evaluating the performance of our proposed approach.

\begin{table}[t]
\centering
\scriptsize
\caption{Depth completion errors by different methods on the \uline{test} set of KITTI depth completion benchmark.}
\label{tab:kitti_overall}
\begin{tabular}{l@{\hspace{-1mm}}ccccc}
\toprule
Methods & RGB info?   & RMSE & MAE    &iRMSE   &iMAE\\ \midrule
SparseConvs \cite{Uhrig2017THREEDV}& $\times$ & 1601.33 & 481.27  & 4.94   & 1.78\\ 
IP-Basic \cite{ku2018defense}  & $\times$ & 1288.46             & 302.60 &3.78   &1.29\\
NConv-CNN \cite{eldesokey2018propagating}  & $\times$ & 1268.22            & 360.28 &4.67 &1.52\\ 
Spade-sD \cite{Jaritz20183DV} & $\times$ & 1035.29 & \uline{\textbf{248.32}} & \uline{\textbf{2.60}} & \uline{\textbf{0.98}} \\ 
Sparse-to-Dense(d) \cite{Ma2018arxiv} & $\times$ & 954.36	& 288.64 & 3.21 & 1.35\\ 
Ours w/o RGB    & $\times$ & {\ul \textbf{937.48}}           & 258.48 & 2.93 & 1.14\\
\midrule
Bilateral NN \cite{barron2016fast} & $\surd$ 
 & 1750.00          & 520.00 &-   &-\\ 
ADNN \cite{chodosh2018deep}  & $\surd$ & 1325.37            & 439.48 &59.39   &3.19\\ 
CSPN \cite{Cheng2018ECCV} & $\surd$ & 1019.64 & 279.46 & 2.93 &  1.15\\ 
Spade-RGBsD \cite{Jaritz20183DV} & $\surd$ & 917.64 & \uline{\textbf{234.81}} & \uline{\textbf{2.17}} & \uline{\textbf{0.95}}\\ 
Sparse-to-Dense(gd) \cite{Ma2018arxiv} & $\surd$ & \uline{\textbf{814.73}}	& 249.95  & 2.80 & 1.21\\ 
Ours w/ RGB     & $\surd$   & 841.78           &253.47 &2.73   &1.13 \\ 
\bottomrule
\end{tabular}
\end{table}

\subsection{KITTI depth completion benchmark}
\label{ssec:kitti}

\subsubsection{Data and evaluation metrics}
\label{sssec:kitti_data_metric}

We first evaluate our proposed approach on the KITTI depth completion benchmark \cite{Uhrig2017THREEDV}. Following the experimental setup in \cite{Uhrig2017THREEDV}, 85,898 depth maps are used for training, 1,000 for validation and 1,000 for test. 
The LIDAR depth maps are aligned with RGB images by projecting the depth map into the image coordinates according to the two sensors' essential matrix. The input depth maps generally contains $<10\%$ sparse points with depth values and the top 1/3 of the input maps do not contain any depth measurements. One example is shown in Fig. \ref{fig:intro}(a).

According to the benchmark, all algorithms are evaluated according to the following metrics, root mean square error (RMSE in mm), mean absolute error (MAE in mm), root mean squared error of the inverse depth (iRMSE in $1/$km), and mean absolute error of the inverse depth (iMAE in $1/$km), i.e.,
\begin{align}
\text{RMSE} &= \left( \frac{1}{\boldsymbol{|V|}} \sum_{u,v \in \boldsymbol{V}} |\boldsymbol{o}(u,v) - \boldsymbol{t}(u,v) |^2 \right)^{0.5}, \label{eq:rmse}\\
\text{MAE} &= \frac{1}{\boldsymbol{|V|}} \sum_{u,v \in \boldsymbol{V}} |\boldsymbol{o}(u,v) - \boldsymbol{t}(u,v) |, \label{eq:mae}\\
\text{iRMSE} &= \left( \frac{1}{\boldsymbol{|V|}} \sum_{u,v \in \boldsymbol{V}} \left| \frac{1}{\boldsymbol{o}(u,v)} - \frac{1}{\boldsymbol{t}(u,v)} \right|^2 \right)^{0.5}, \label{eq:irmse}\\
\text{iMAE} &= \frac{1}{\boldsymbol{|V|}} \sum_{u,v \in \boldsymbol{V}} \left| \frac{1}{\boldsymbol{o}(u,v)} - \frac{1}{\boldsymbol{t}(u,v)} \right|, \label{eq:imae}
\end{align}
where $\boldsymbol{o}$ and $\boldsymbol{t}$ represent the output of our approach and ground-truth depth values.

For RMSE and MAE, RMSE is more sensitive to large errors compared. This is because even a small number of large errors would be magnified by the square operation and dominate the overall loss value. RMSE is therefore chosen as the main metric for ranking different algorithms in the KITTI leaderboard. Since large depth values usually have greater errors and might dominate the calculation of RMSE and MAE, iRMSE and iMAE are also evaluated, which calculate the mean of inverse of depth errors. In this way, large depth values' errors would have much lower weights on the two metrics. The two metrics focus more on depth points near the LIDAR sensors.

\subsubsection{Comparison with state-of-the-arts}

The performance of our proposed approaches and state-of-the-art depth completion methods are recorded in Table \ref{tab:kitti_overall}.

SparseConvs represents the 6-layer convolution neural network with only sparsity-invariant convolution proposed in \cite{Uhrig2017THREEDV}. It only supports convolution and max-pooling operations and therefore loses much high-resolution information. IP-Basic represents the method in \cite{ku2018defense}, a well-designed algorithm hand-crafted rules based on several traditional image processing algorithms. NConv-CNN \cite{eldesokey2018propagating} proposed a constrained convolution layer and propagating confidence across layers for depth completion. Bilateral NN \cite{barron2016fast} uses RGB images as guidance and integrate bilaterial filters into deep neural networks. It was modified to handle sparse depth completion following \cite{Uhrig2017THREEDV}. Spade-sD and Spade-RGBsD \cite{Jaritz20183DV} do not have special treatment for sparse data. They utilize conventional dense CNN but adopt different loss function and training strategy. CSPN \cite{Cheng2018ECCV} iteratively learns inter-pixel affinities with RGB guidance via recurrent convolution operations. The affinities could then be used to spatially propagate depth values between different locations. Sparse-to-dense(d) and Sparse-to-dense(gd) \cite{Ma2018arxiv} explore additional temporal information from sequential data to apply additional supervisions based on the photometric loss between neighboring frames.

For methods without RGB guidance, our proposed network without RGB guidance outperforms all other peer-reviewed methods in terms of RMSE (the main ranking metric in KITTI leaderboard). Spade-sD has better MAE, iRMSE and iMAE, which mean that this method performs better on nearby objects but is more likely to generate large errors than our proposed method. Note that we utilize the $L2$ loss function to deliberately minimize RMSE. If other metrics are considered to be more important, different loss functions could be adopted for training.

For methods with RGB guidance, our method ranks 2nd behind Sparse-to-dense(gd) \cite{Ma2018arxiv} in terms of RMSE. However, Sparse-to-dense(gd) utilized additional supervisions from temporal information, while our proposed method only uses supervisions from individual frames. 

\begin{table}[t]
\centering
\caption{Component analysis of our proposed method on the \uline{validation} set of KITTI depth completion benchmark.}
\label{tab:components}
\begin{tabular}{lcccc}
\toprule
Method      & RMSE & MAE\\ 
\midrule
Baseline w/o sparseconv   & 1819.81             & 426.84\\ 
{Baseline w/ sparseconv}  & {1683.22}             & {447.93}\\ 
{Baseline + MS (Up only)}   & {1185.02}          & {323.41}\\ 
{Baseline + MS (Down only)}  & {1192.43}          & {322.95}\\ 
{Baseline + MS (Mid-level flow removed)}  & {1166.87}          & {317.74}\\ 
Baseline + MS (Full)  & 1137.42          & 315.32\\ 
Baseline + MS + SO \ \ \ \ \ & \ \ \ \ \ 994.14 \ \ \ \ \             & \ \ \ \ \ 262.41 \ \ \ \ \ \\ 
Baseline + MS + SO + RGB    & 883.74             & 257.11 \\ 
\bottomrule
\end{tabular}
\end{table}

\subsubsection{Ablation study}

{We investigate individual components in our framework to see whether they contribute to the final performance. The investigated components include, multi-scale structure, sparsity-invariant operations, and RGB guidance. We choose our full-resolution low-level feature path without the mid-level or high-level flow in our network (i.e., the upper path in Fig. \ref{fig:overall}(a)) as the baseline model for this section. The baseline model does not include RGB guidance. The analysis results on KITTI validation set are shown in Table \ref{tab:components}. And the baseline model without multi-scale feature fusion generates large depth estimation errors.} 

\begin{figure*}
\centering
\begin{tabular}{c c c}
\includegraphics[height=1.75cm]{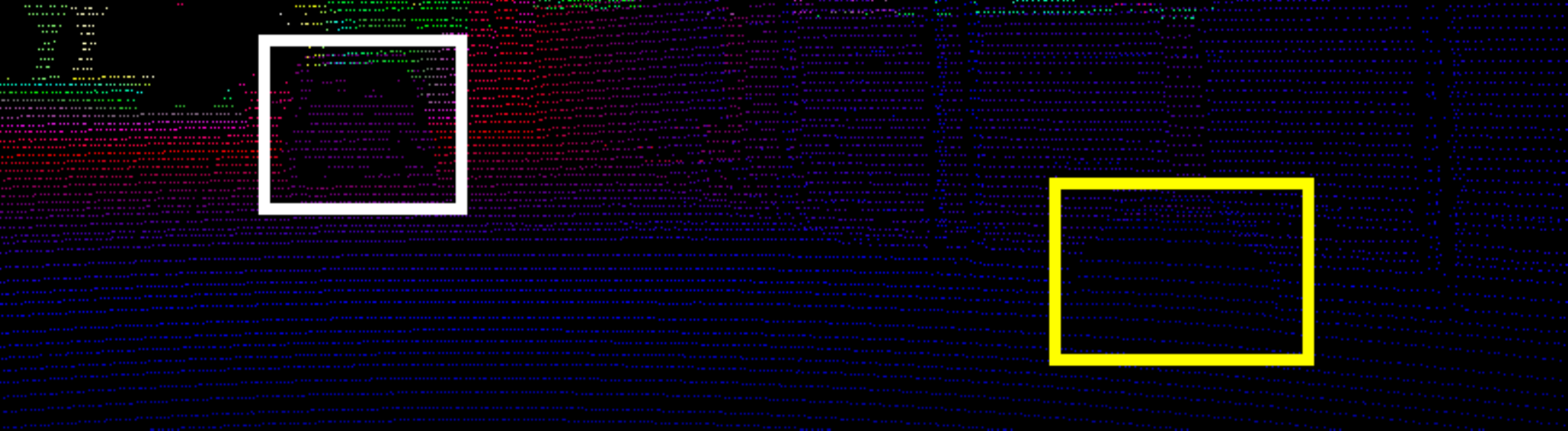} & \includegraphics[height=1.75cm]{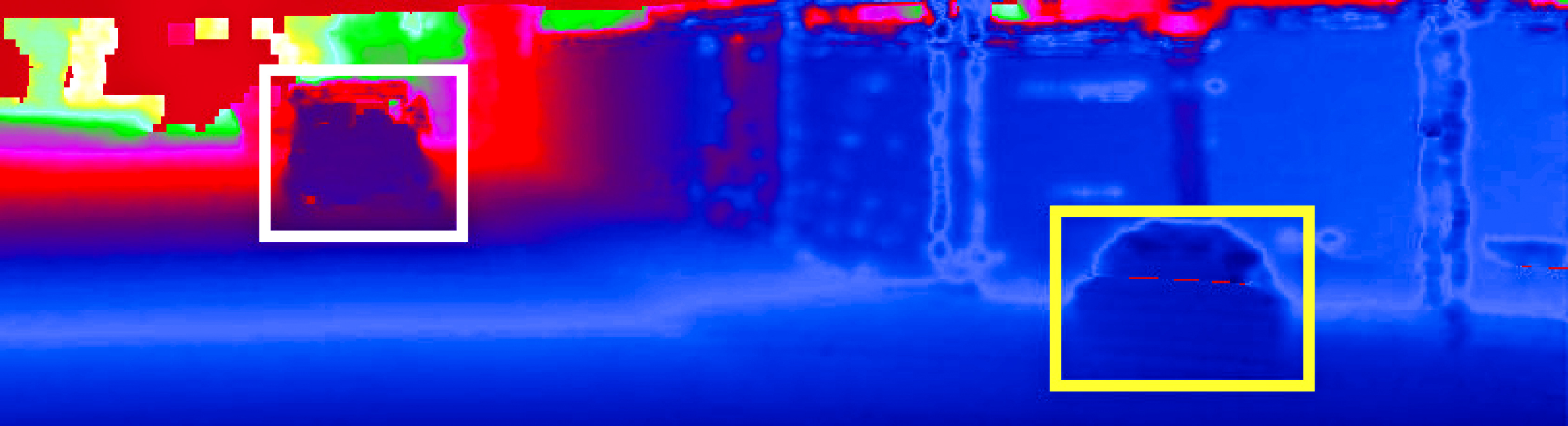}\\ 
Input Sparse Depth Map & Baseline\\ 
\includegraphics[height=1.75cm]{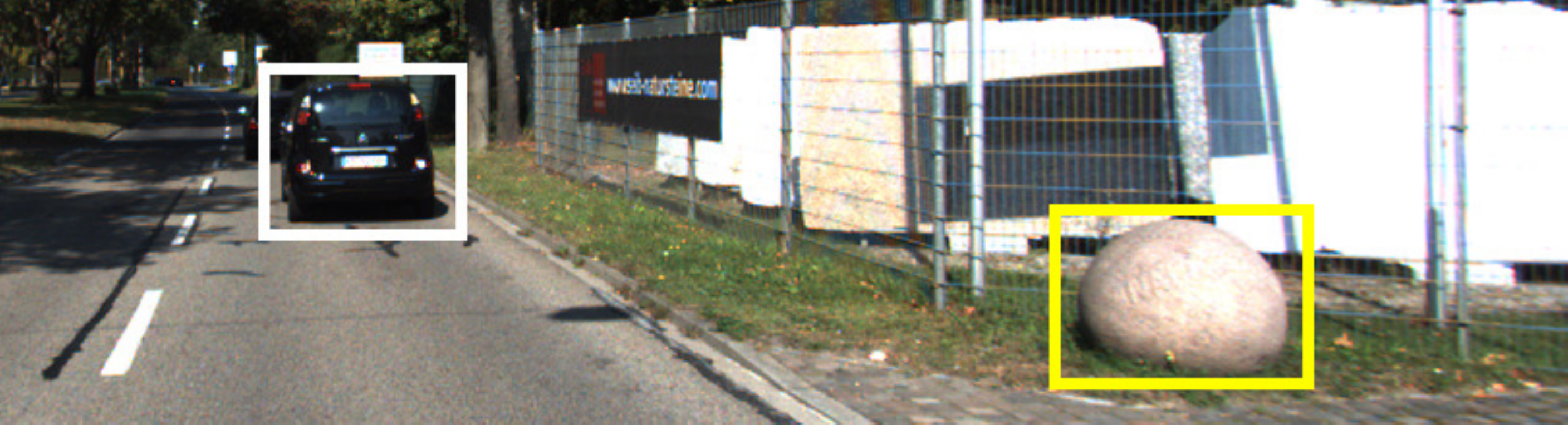} & \includegraphics[height=1.75cm]{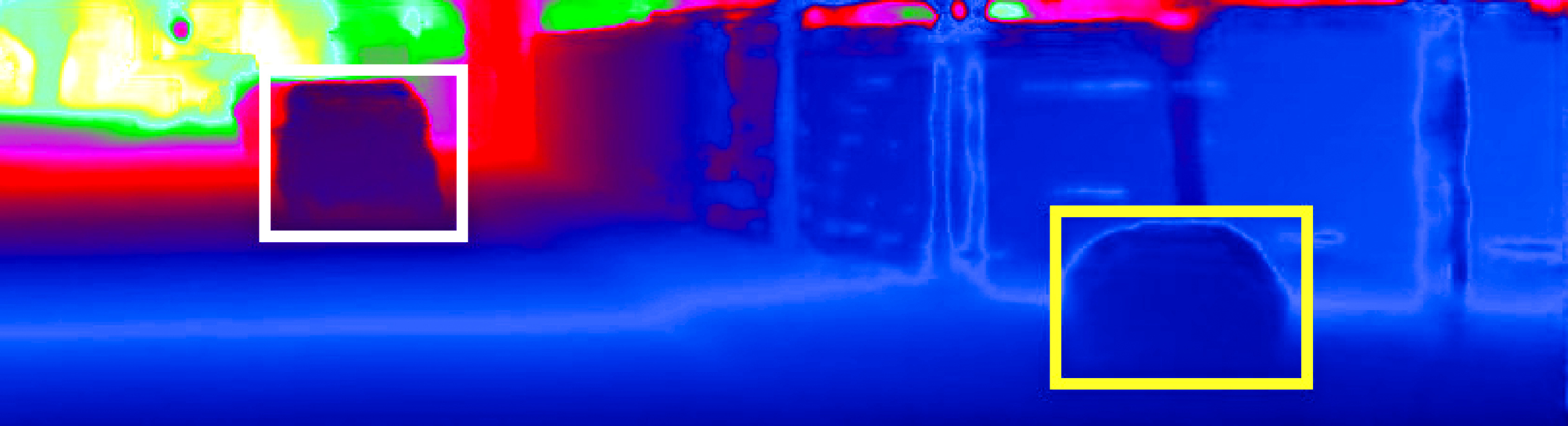}\\ 
Corresponding Image & Baseline + MS\\ 
\end{tabular}
\caption{An example in the KITTI validation set to show the results by the baseline model (\emph{Baseline}) vs. baseline model with mutli-scale encoder-decoder structure (\emph{Baseline + MS}). See rectangles for better boundary regions by \emph{Baseline + MS}.}
\label{fig:example_ms}
\end{figure*}

\textbf{Multi-scale structure.} {Using our multi-scale encoder-decoder structure (denoted as \emph{Baseline+MS (Full)}) in addition to the baseline model enables fusing low-level and high-level information from different scales. The multi-scale structure provides much larger receptive fields for neurons in the last convolution layer. Therefore, even if some regions in the input depth map are very sparse, the model could still predict depth values for every grid points. Using multi-scale features generally results in clearer boundaries and shows higher robustness to noises. \emph{Baseline+MS (Full)} significantly decreases RMSE from 1819.81 to 1137.42. Also, our fusing skip connections connect different scales also enable a better fusion for the information in our network. By removing either the up fusing skip connection (shown by up arrows within the blocks in Fig. 6(a)) or the down fusing skip connection, our network performs worse as shown in Table 2 (Up only and Down only entries). Also, the mid-level flow is making the performance better (Mid-level flow removed entry). An example showing the differences between \emph{Baseline+MS (Full)} and \emph{Baseline} is in Fig.~\ref{fig:example_ms}.}

\begin{figure*}
\centering
\begin{tabular}{c c c}
\includegraphics[height=1.6cm]{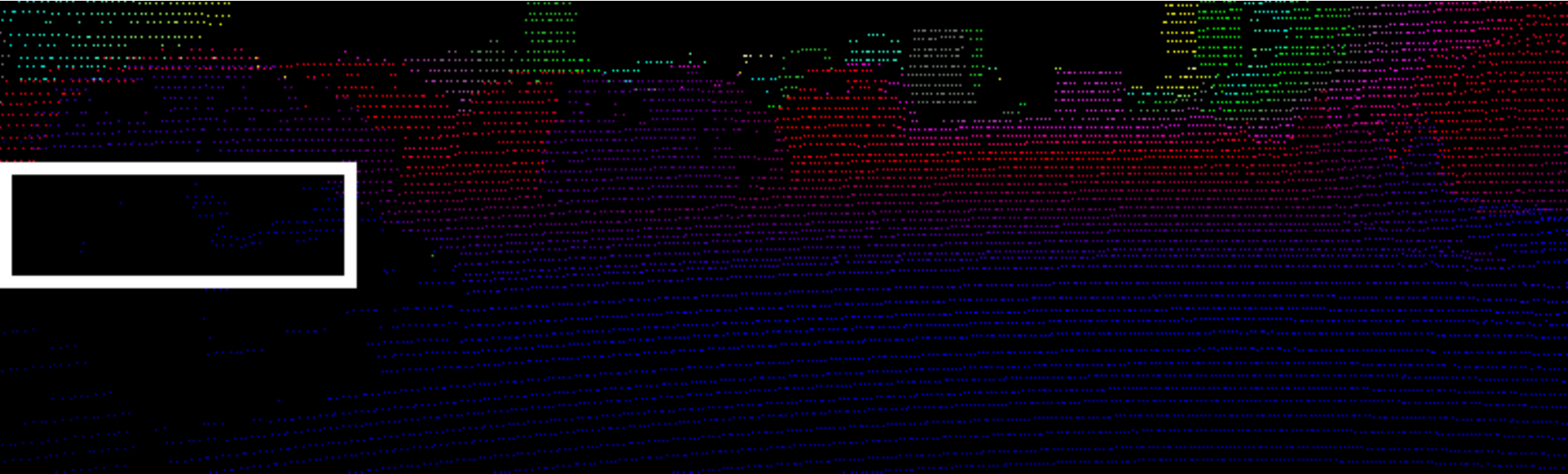} & \includegraphics[height=1.6cm]{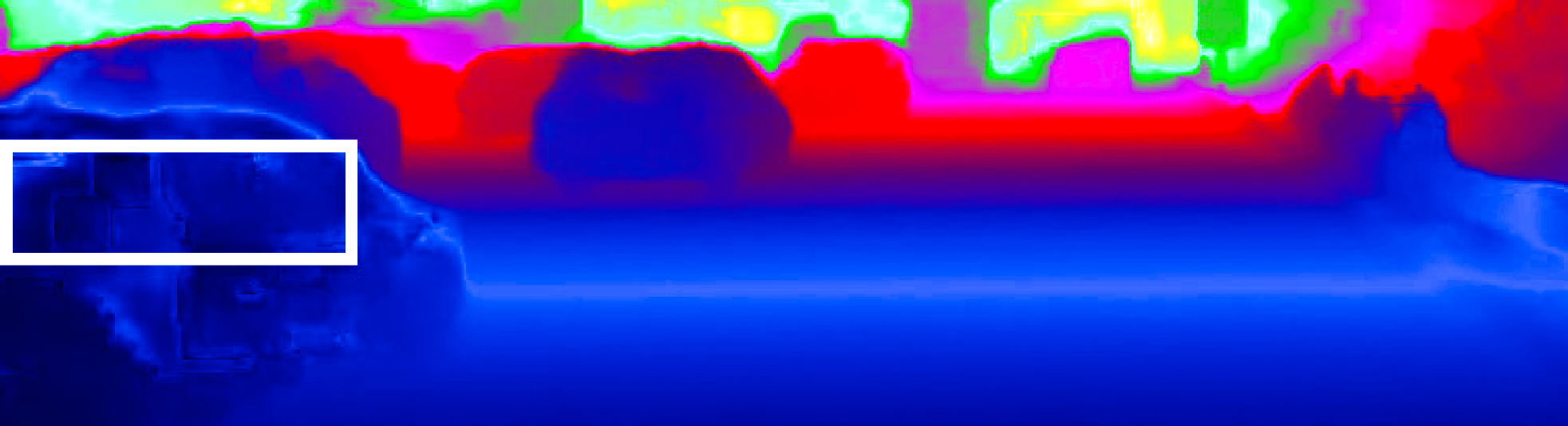} & \includegraphics[height=1.6cm]{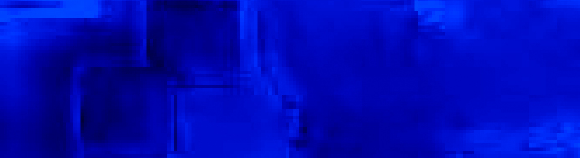} \\ 
Input Sparse Depth Map & Baseline + MS & Baseline + MS (magnified) \\ 
\includegraphics[height=1.6cm]{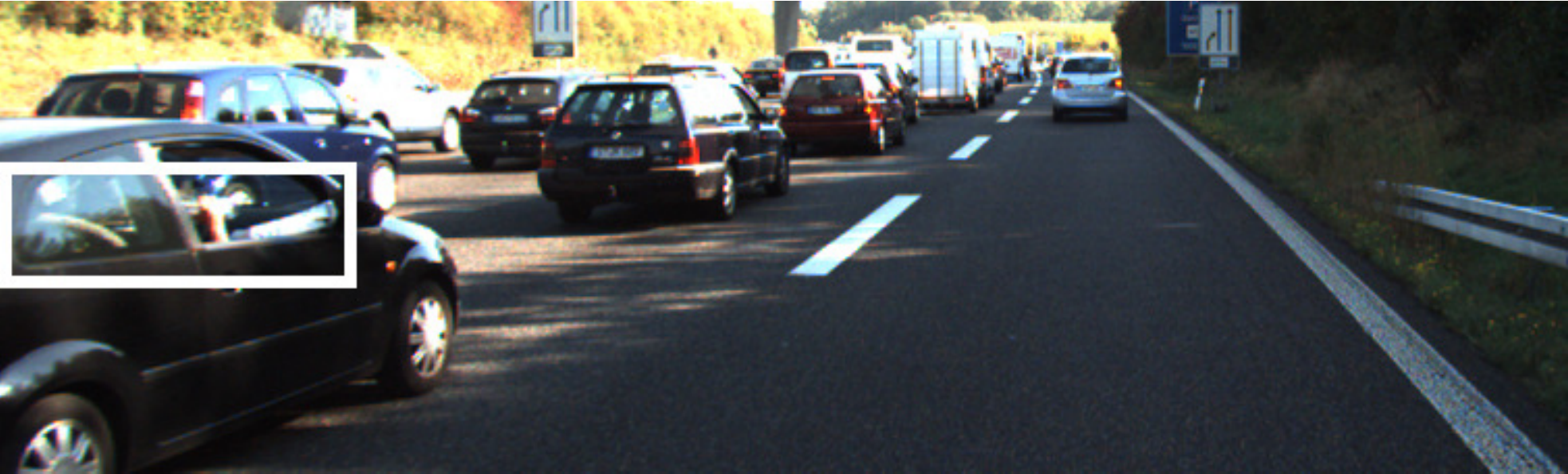} & \includegraphics[height=1.6cm]{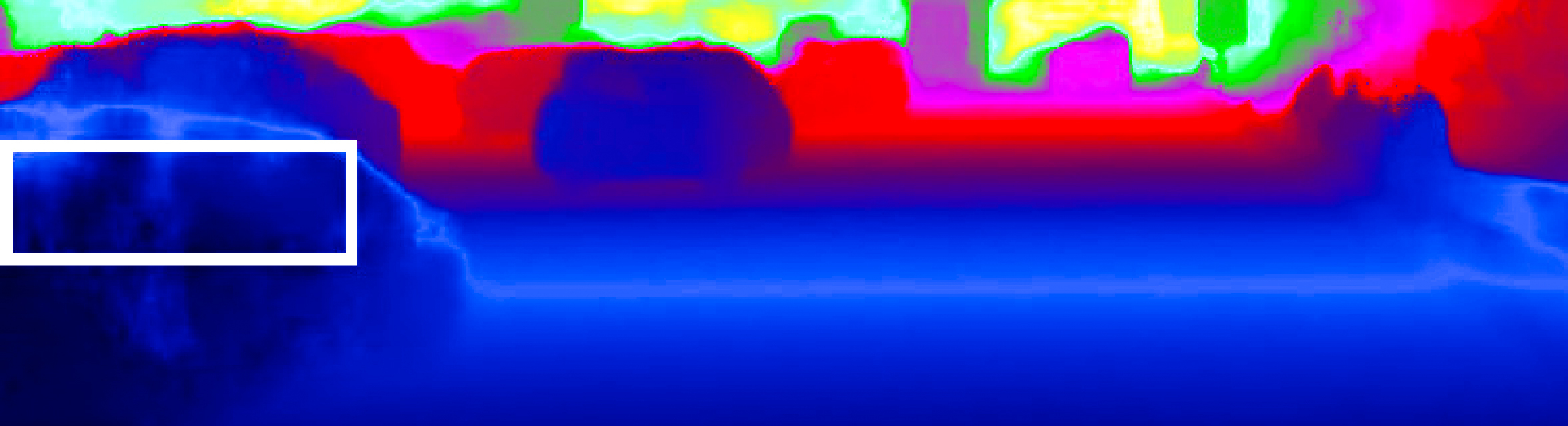} & \includegraphics[height=1.6cm]{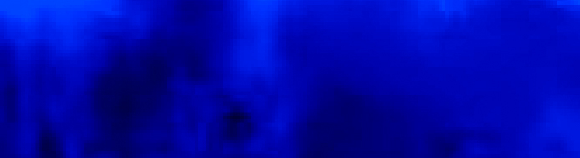} \\ 
Corresponding Image & Baseline + MS + SO & Baseline + MS + SO (magnified) \\ 
\end{tabular}
\caption{An example in the KITTI validation set to show the effectiveness with (\emph{Baseline + MS}) and without proposed sparsity-invariant operations (\emph{Baseline + MS + SO}). Rectangles show \emph{Baseline + MS + SO} better handles input depth maps with very sparse depth points.}
\label{fig:example_so}
\end{figure*}

\textbf{Sparsity-invariant operations.} The \emph{Baseline+MS} utilizes dense convolution. It has difficulty in handling sparse input data and sparse feature maps, especially for regions where there are very sparse points. The \emph{Baseline+MS+SO} uses our proposed sparsity-invariant operations to maintain a correct mask flow and then converts conventional operations in \emph{Baseline+MS} into sparsity-invariant ones to handle very sparse inputs. RMSE by \emph{Baseline+MS+SO} improves from 1137.42 to 994.14. An example is shown in Fig. \ref{fig:example_so}, which shows \emph{Baseline+MS+SO} better handles regions with very sparse inputs.

\begin{figure*}[t]
\centering
\begin{tabular}{c c c}
\includegraphics[height=1.75cm]{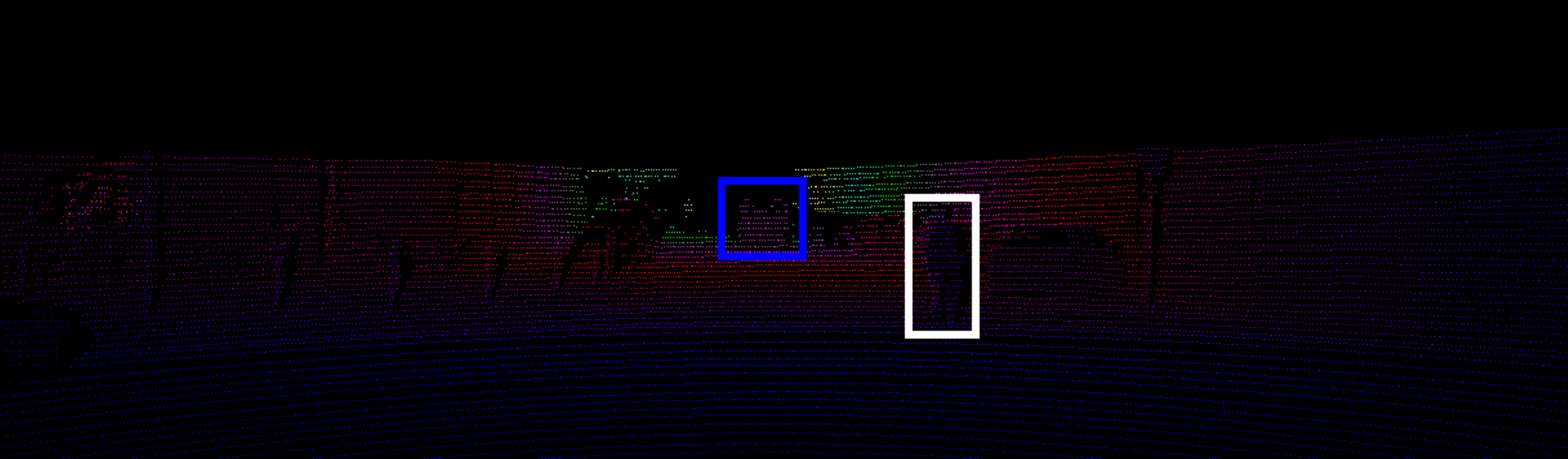} & \includegraphics[height=1.75cm]{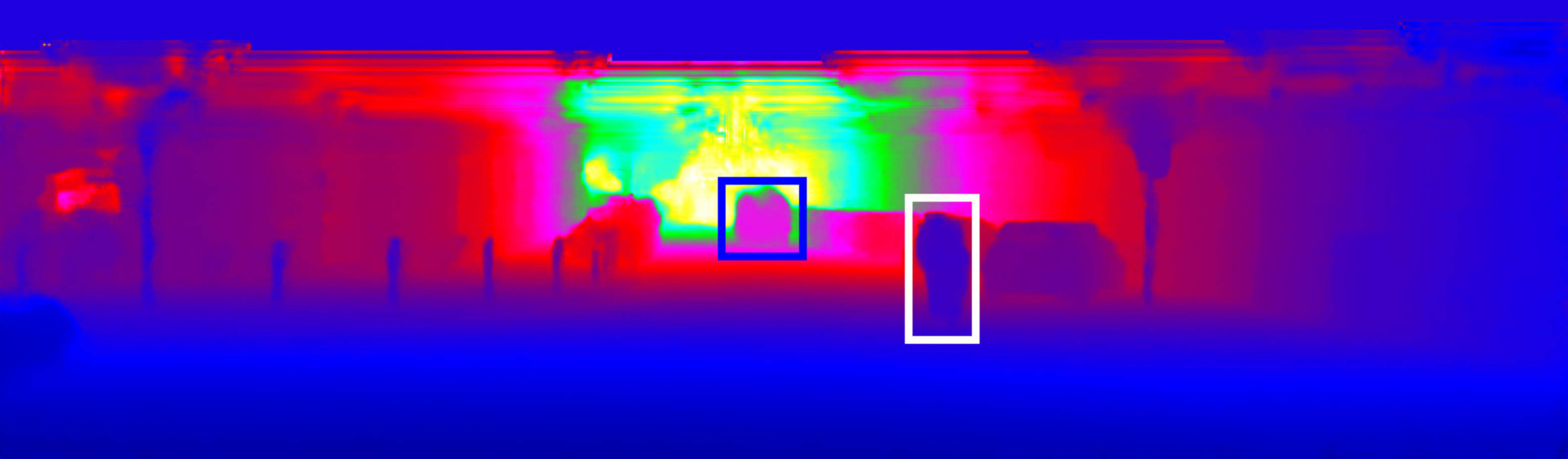}\\ 
Input Sparse Depth Map & Baseline + MS + SO\\ 
\includegraphics[height=1.75cm]{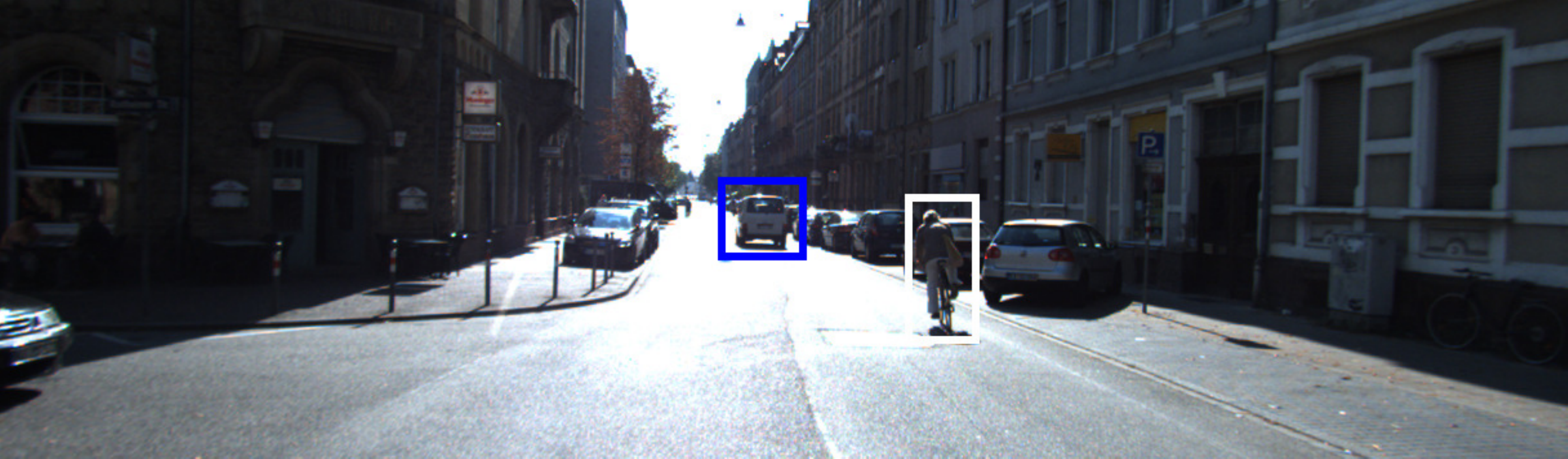} & \includegraphics[height=1.75cm]{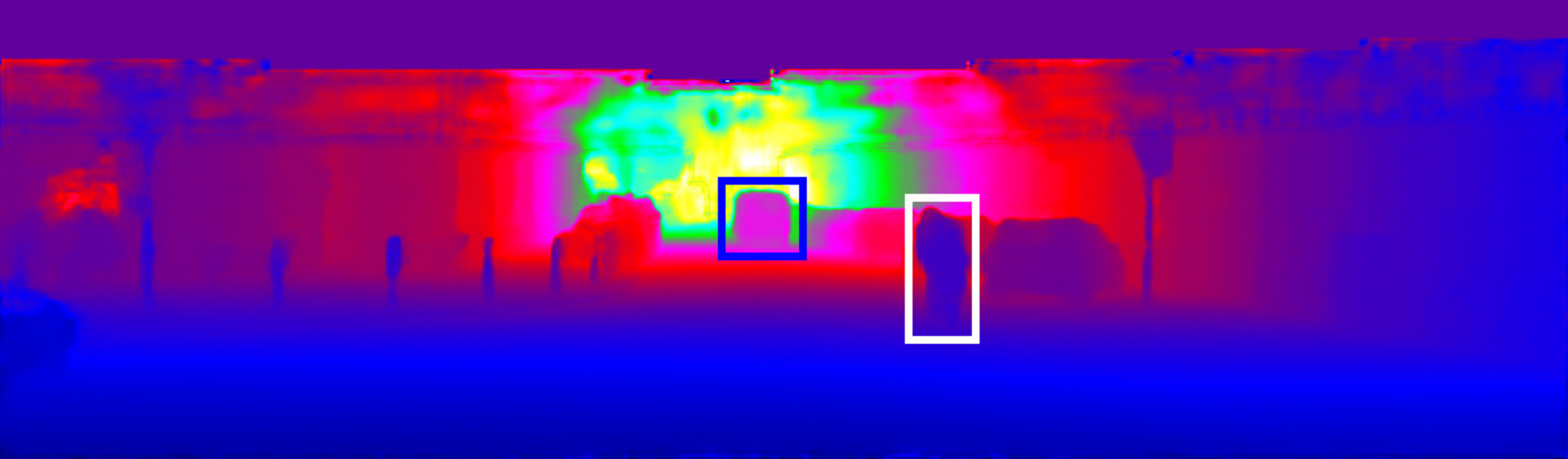}\\ 
Corresponding Image & Baseline + MS + SO + RGB\\ 
\end{tabular}
\caption{An example in the KITTI validation set to show the effectiveness withour RGB guidance (\emph{Baseline + MS + SO}) and with RGB guidance (\emph{Baseline + MS + SO + RGB}). Rectangles show \emph{Baseline + MS + SO + RGB} generates clearer boundaries of the cyclist and the far-away vehicle.}
\label{fig:examples_rgb}
\end{figure*}

\begin{table}[t]
		\centering
		\label{tab:structures}
				\caption{Comparison with commonly used encoder-decoder networks without RGB guidance on KITTI depth completion \uline{validation} set.}
		\scriptsize
		\begin{tabular}{ccccc}
			\toprule
			Method      & {\textbf{RMSE}} & MAE\\ \midrule
			U-Net \cite{ronneberger2015u}   & 1387.35            & 445.73\\ 
			FRRN \cite{pohlen2017full}   & 1148.27            & 338.56\\ 
			PSPNet \cite{zhao2017pyramid}   & 1185.39             & 354.21\\ 
			FPN \cite{lin2017feature} &1441.82 & 473.65 \\
			{He et al. \cite{he2018learning}} & {1056.39}	& {293.86}\\ 
			Ours w/o RGB       & {\ul \textbf{994.14}}           & {\ul \textbf{262.41}}\\ \bottomrule
		\end{tabular}
\end{table}

\textbf{Deep fusion of RGB features.} By incorporating RGB features into \emph{Baseline+MS+SO+RGB}, the network utilizes useful additional guidance from RGB images for further improving the depth completion accuracy. An example comparing \emph{Baseline+MS+SO+RGB} and \emph{Baseline+MS+SO} is shown in Fig. \ref{fig:examples_rgb}, where the resulting depth maps with RGB guidance are much sharper at image boundaries.

\subsubsection{Comparison with other encoder-decoder structures}

{To evaluate the effectiveness of our proposed HMS-Net structure, we conduct experiments to test other commonly used encoder-decoder network structures. We modify those structures with the sparsity-invariant convolution and our proposed sparsity-invariant operations to handle sparse inputs. The experimental results on the KITTI validation set are reported in Table 3. We compared our network structure without RGB guidance with modified U-Net \cite{ronneberger2015u}, FRRN \cite{pohlen2017full}, PSPNet \cite{zhao2017pyramid}, FPN \cite{lin2017feature} and He et al. \cite{he2018learning} without focal length. Our proposed network structure achieves the lowest errors in terms of RMSE and MAE.}

\begin{figure*}
	\centering
	\footnotesize
	\begin{tabular}{c@{\hspace{-3.5mm}}c@{\hspace{-2.5mm}}c@{\hspace{-2.5mm}}c}
		& \includegraphics[height=4.3cm]{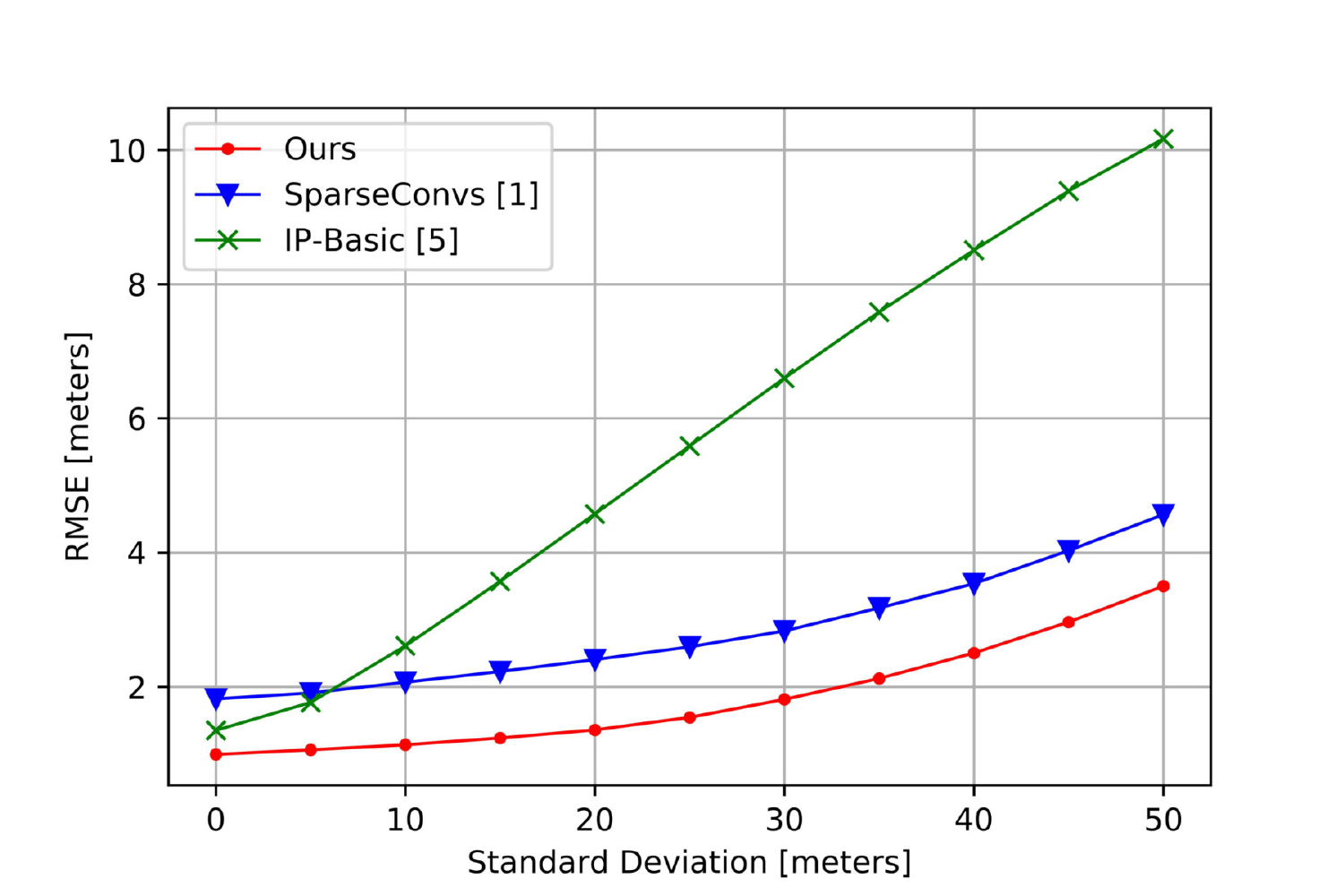} & \includegraphics[height=4.3cm]{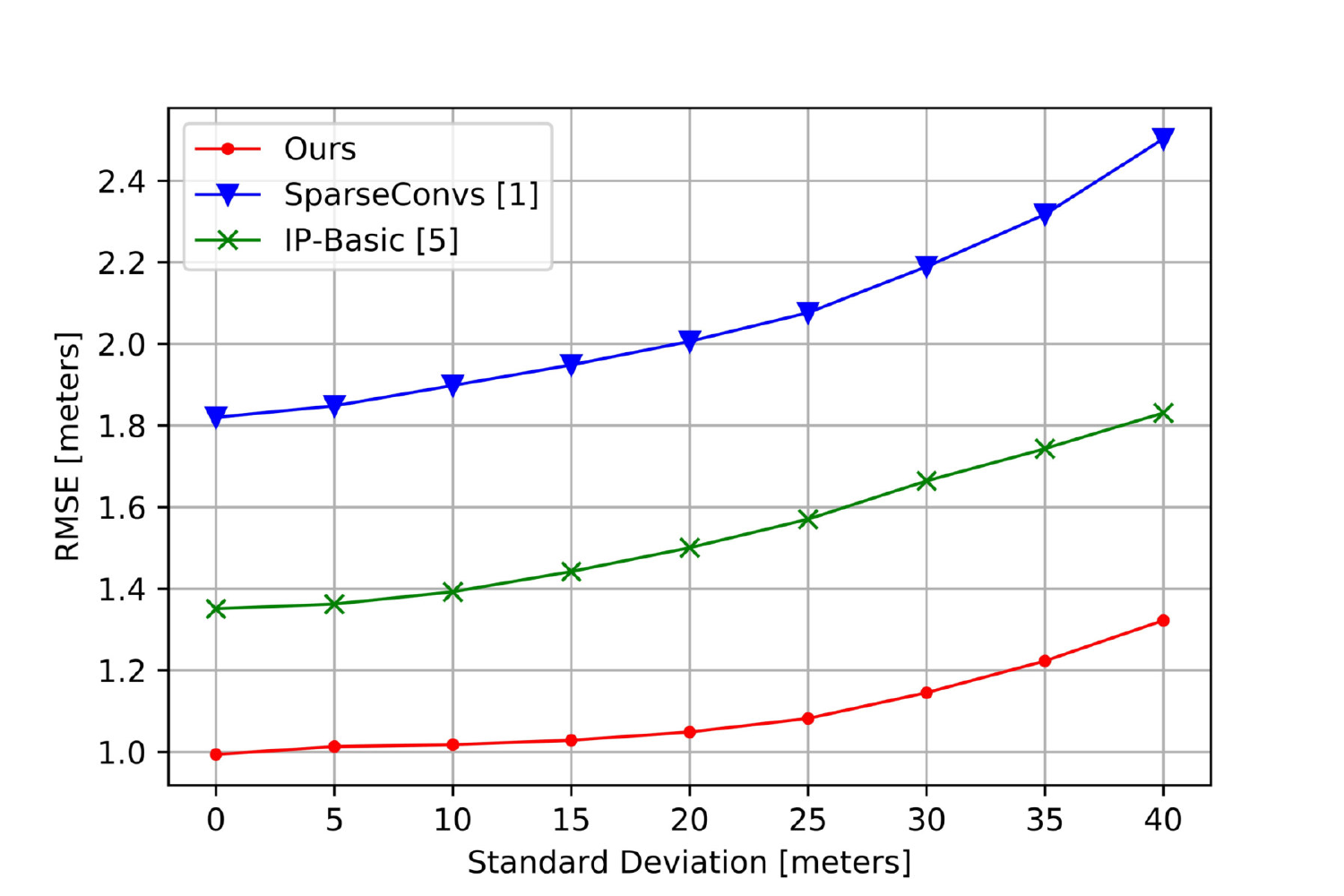} & \includegraphics[height=4.3cm]{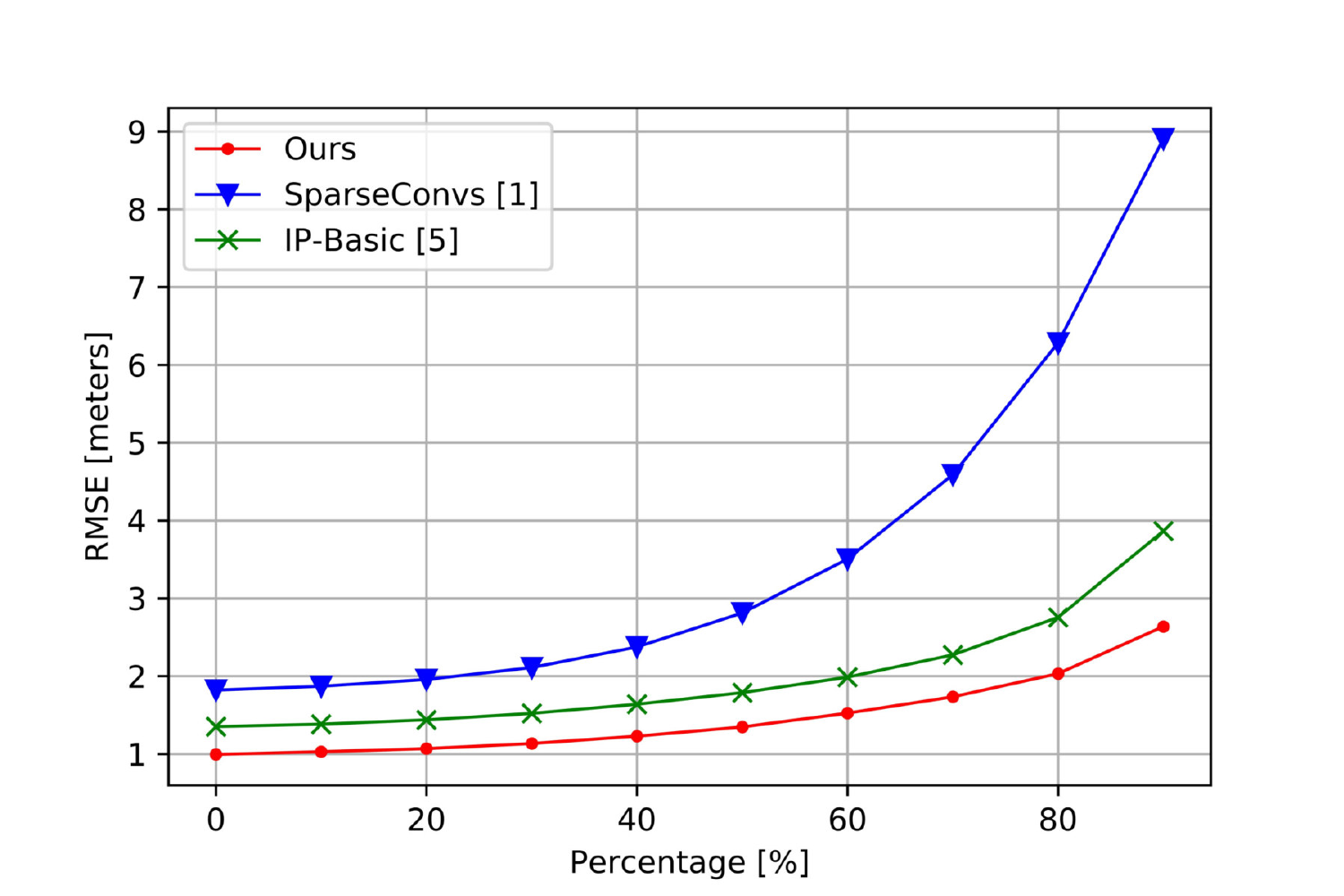}\\
	& (a) Scene-level Gaussian noises & (b) Region-level Gaussian noise & (c) Randomly abandon input depth points
	\end{tabular}
	\caption{Testing robustness on the validation set of KITTI depth completion benchmark with manually corrupted data. (a) Scene-level Gaussian noises on 10\% depth points. (b) Region-level Gaussian noise on eight 25$\times$ 25 regions. (c) Randomly abandon input depth points with varying probabilities.}
	\label{fig:robustness}
\end{figure*}

\begin{table*}[t]
	\centering
	\caption{Comparison with other methods on NYU-depth-v2 depth dataset with varying $N$ values.}
	\begin{tabular}{ccccccccccc}
	\toprule
	\multirow{2}{*}{Method} & \multicolumn{2}{c}{20 samples} & \multicolumn{2}{c}{50 samples} & \multicolumn{2}{c}{200 samples}\\ 
	& RMSE          & REL         & RMSE          & REL         & RMSE          & REL        \\ \hline
	Ma et al. \cite{ma2017sparse} w/o RGB                  & 0.461          & \uline{\textbf{0.110}}         & 0.347          & 0.076         & 0.259          & 0.054       \\ 
	{Jaritz et al. \cite{Jaritz20183DV} w/o RGB}                  & {0.476}          & {0.114}         & {0.358}          & {0.074}         & {0.246}          & {0.051}       \\ 
	{Ma et al. \cite{Ma2018arxiv} w/o RGB}                  & {0.481}          & {0.113}         & {0.352}          & {\uline{\textbf{0.073}}}         & {0.245}          & {0.049}       \\ 
	{He et al. \cite{he2018learning} w/o RGB}   & {0.478}          & {0.113}         & {0.355}          & {0.072}         & {0.238}          & {0.045}       \\
	Ours w/o RGB                 & \uline{\textbf{0.449}}          & \uline{\textbf{0.110}}         & \uline{\textbf{0.344}}          & \uline{\textbf{0.073}}        & \uline{\textbf{0.233}}         & \uline{\textbf{0.044}}  \\
	\midrule
	{Ma et al. \cite{ma2017sparse} w/ RGB}                  & {0.351}          & {\uline{\textbf{0.078}}}         & {0.281}          & {\uline{\textbf{0.059}}}         & {0.230}          & {0.044}       \\ 
	{Ours w/ RGB}                   & {\uline{\textbf{0.350}}}          & {0.079}         & {\uline{\textbf{0.274}}}          & {\uline{\textbf{0.059}}}         & {\uline{\textbf{0.212}}}          & {\uline{\textbf{0.041}}}       \\ 
	\bottomrule
\end{tabular}
\label{tab:nyuv2}
\end{table*}

\subsection{Robustness testing on KITTI benchmark}
\label{ssec:robustness}

\subsubsection{Robustness to depth noises}

Since the sparse depth maps are obtained by LIDAR scans, inevitably, there would be noises in acquired depth values. As a result, the robustness of depth completion algorithms with regard to different noise levels is important in practice. We conduct experiments to test the robustness of our proposed model without RGB guidance and compare with SparseConvs \cite{Uhrig2017THREEDV} and IP-Basic \cite{ku2018defense}. Note that all models in this section are trained on original data and directly tested on noisy data.

\textbf{Scene-level Gaussian noises.} For this experiment, we add Gaussian noises on randomly selected 10\% depth points among all points. Once the 10\% points are selected, for a specific noise standard deviation level from 5-50 meters, we sample additive noise values from the zero-mean Gaussian distribution. The negative additive noises could simulate occlusion from raindrops, snowflakes or fog, while the positive additive noises mimic the laser going through glasses that mistakenly measures objects behind glasses. Noisy points whose depth values are smaller than 1 meter are set to 1 meter to simulate the minimum range of the LIDAR sensor. The RMSEs by three methods on the KITTI validation set with Gaussian noises are shown in Fig. \ref{fig:robustness}(a). Our method outperforms both SparseConvs \cite{Uhrig2017THREEDV} and IP-Basic \cite{ku2018defense} on different noisy depth values.

\textbf{Region-level Gaussian noises.} We randomly select eight regions of size 25 $\times$ 25 pixels in every input depth map. In each region, 50\% of depth points are randomly selected to add Gaussian noises of zero mean and different standard deviation values. Noisy points whose depth values are smaller than 1 meter are set to 1 meter to simulate the minimum range of the LIDAR sensor. The region-level noises are used to simulate the cases where large glasses or mirrors exist. Those regions would reflect most laser and leave large holes in the obtained depth map. The RMSE by different methods on the KITTI validation set are shown in Fig. \ref{fig:robustness}(b). Our method again outperforms the two compared methods, because of its capability of fusing low-level and high-level information with the proposed multi-scale encoder-decoder structure.

\subsubsection{Robustness to sparsity}

Robustness to sparsity is also essential to depth completion algorithms. We conduct experiments on testing different levels of sparsity of the input depth maps. For each input depth map, we randomly abondon 10\%-90\% of valid depth points. Note again that all methods are trained on original training data and are not finetuned to adapt the sparser inputs. The results on the KITTI validation set by the our model without RGB guidance and two compared methods are shown in Fig. \ref{fig:robustness}(c). Our proposed method shows the highest tolerance against different sparsity levels.

\begin{figure*}[h!]
	\centering
	\footnotesize
	\begin{tabular}{l@{\hspace{-2mm}}m{1cm}>{\centering\arraybackslash}m{3cm}>{\centering\arraybackslash}m{3cm}>{\centering\arraybackslash}m{3cm}>{\centering\arraybackslash}m{3cm}>{\centering\arraybackslash}m{3cm}}
		& Inputs 
		& \includegraphics[height=2.35cm]{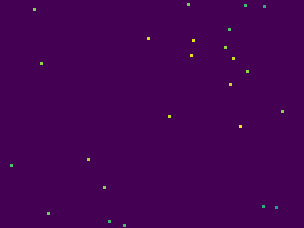}
		& \includegraphics[height=2.35cm]{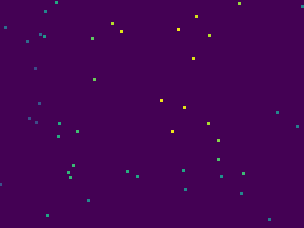}
		& \includegraphics[height=2.35cm]{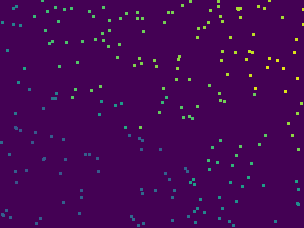}
		& \includegraphics[height=2.35cm]{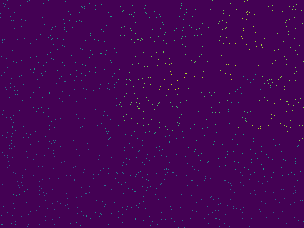}
		& \includegraphics[height=2.35cm]{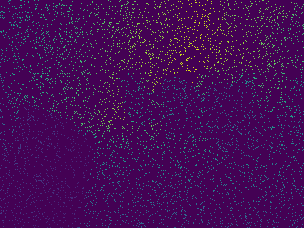}\\
		& RGB Images (not used)
		& \includegraphics[height=2.35cm]{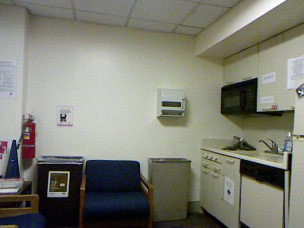}
		& \includegraphics[height=2.35cm]{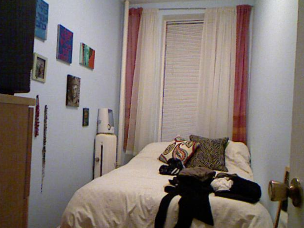}
		& \includegraphics[height=2.35cm]{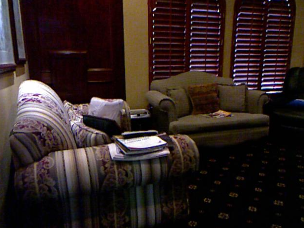}
		& \includegraphics[height=2.35cm]{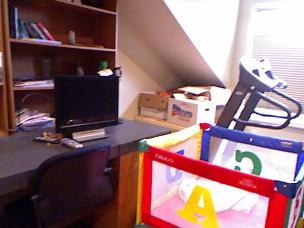}
		& \includegraphics[height=2.35cm]{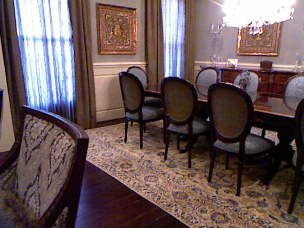}\\
		& Prediction Results
		& \includegraphics[height=2.35cm]{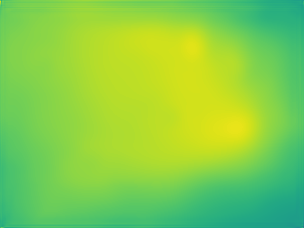}
		& \includegraphics[height=2.35cm]{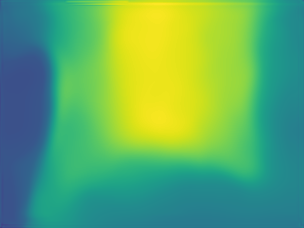}
		& \includegraphics[height=2.35cm]{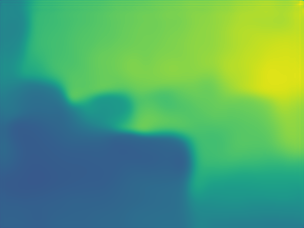}
		& \includegraphics[height=2.35cm]{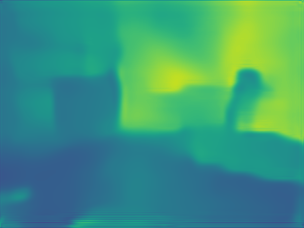}
		& \includegraphics[height=2.35cm]{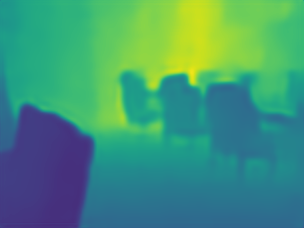}\\
		& Ground Truth
		& \includegraphics[height=2.35cm]{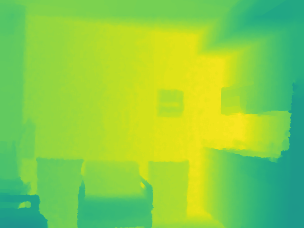}
		& \includegraphics[height=2.35cm]{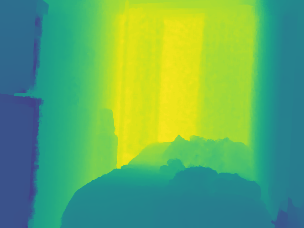}
		& \includegraphics[height=2.35cm]{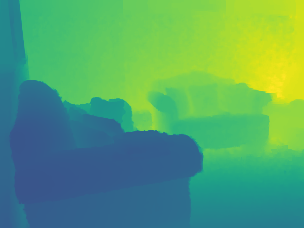}
		& \includegraphics[height=2.35cm]{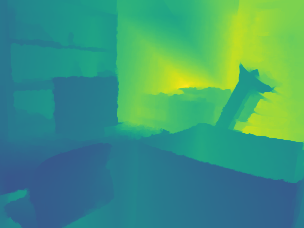}
		& \includegraphics[height=2.35cm]{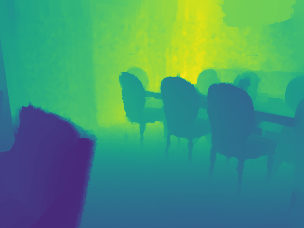}\\
		& & $N=20$ & $N=50$ & $N=200$ & $N=1000$ & $N=5000$
	\end{tabular}
	\caption{Depth completion examples in NYU-depth-v2 dataset \cite{silberman2012indoor} by our proposed method with vary $N$ values. (First row) Input sparse depth maps. (Second row) Corresponding RGB images (not used as algorithm inputs). (Third row) Predicted dense depth maps by our proposed method. (Fourth row) Ground truth dense depth maps.}
	\label{fig:nyu}
\end{figure*}

\subsection{NYU-depth-v2 dataset}

\subsubsection{Data, experimental setup, and evaluation metrics}

We also evaluate our proposed method on the NYU-depth-v2 dataset \cite{silberman2012indoor} with its official train/test split. Each RGB image in the dataset is paired with a spatially aligned dense depth map. The original depth maps are dense and the dataset is not originally proposed for sparse depth completion. Following the experimental setup in \cite{ma2017sparse}, synthetic sparse depth maps could be created to test the performance of depth completion algorithms. Only $N$ depth points in each depth map are randomly kept as input depth maps for depth completion. The training set consists of depth and image pairs from 249 scenes, and 654 images are selected for evaluating the final performance according to the setup in \cite{ma2017sparse,laina2016deeper,eigen2014depth}. RMSE (Eq. \eqref{eq:rmse}) and mean absolute relative error (REL in meters) are adopted as the evaluation metrics. REL is calculated as
\begin{align}
\text{REL} = \frac{1}{\boldsymbol{|V|}} \sum_{u,v \in \boldsymbol{V}} \left|\frac{\boldsymbol{o}(u,v) - \boldsymbol{t}(u,v)}{\boldsymbol{t}(u,v) }\right|, \label{eq:rel}
\end{align}
where $\boldsymbol{o}$ and $\boldsymbol{t}$ are the outputs of our network and the ground truth dense depth maps.

\subsubsection{Comparison with state-of-the-art}

{We compare our method to methods proposed in Ma et al. \cite{Ma2018arxiv,ma2017sparse}\footnote{The code released by the authors were utilized.}, Jaritz et al. \cite{Jaritz20183DV} and He et al.\cite{he2018learning}.}
Since the input depth maps are much sparser than the depth maps in KITTI dataset \cite{Uhrig2017THREEDV}, we added a $2 \times 2$ max-pooling layer following the first $5 \times 5$ convolution layer, and added a Batch Normalization \cite{ioffe2015batch} layer after each convolution layer. Each input depth map has $N= 20$, $50$, $200$ randomly kept depth points. RMSE and REL of different $N$ values are reported in Table \ref{tab:nyuv2}, which demonstrates that our proposed method outperforms all other methods without using RGB information. {We also show the result with RGB guidance in TABLE 4 as well as examples of different $N$ values and the depth completion results in Fig. \ref{fig:nyu}.}

\section{Conclusions}

In this paper, we proposed several novel sparsity-invariant operations for handling sparse feature maps. The novel operations enable us to design a novel sparsity-invariant encoder-decoder network, which effectively fuses multi-scale features from different CNN layers for accurate depth completion. RGB features for better guiding depth completion is also integrated into the proposed framework. Extensive experiment results and component analysis show advantages of the proposed sparsity-invariant operations and the encoder-decoder network structure. The proposed method outperforms state-of-the-arts and demonstrate great robustness against different levels of data corruption.


%


\ifCLASSOPTIONcaptionsoff
  \newpage
  
\fi

\bibliographystyle{bibtex/IEEEtran}
\bibliography{egbib}
\end{document}